\documentclass[a4paper,fleqn]{cas-sc}
\pdfoutput=1
\usepackage[authoryear,longnamesfirst]{natbib}
\usepackage{hyperref}
\usepackage{hhline}

\usepackage{makecell}

\def\tsc#1{\csdef{#1}{\textsc{\lowercase{#1}}\xspace}}
\tsc{WGM}
\tsc{QE}
\begin{document}
\let\WriteBookmarks\relax
\def\floatpagepagefraction{1}
\def\textpagefraction{.001}

\shorttitle{Understanding and Estimating Domain Complexity Across Domains}    
%\shorttitle{Toward Defining an Across-domain Domain Complexity Measure for Action Domains} 
\shortauthors{Doctor, Kejriwal, Holder, Kildebeck, Resmini, Pereyda}  

\title [mode = title]{Understanding and Estimating Domain Complexity Across Domains}  
%\title [mode = title]{Toward Defining an Across-domain Domain Complexity Measure for Action Domains}  

%\tnotemark[<tnote number>] 

%\tnotetext[<tnote number>]{<tnote text>} 


\author[1]{Katarina Doctor}[orcid=0000-0001-8534-2808]

\cormark[1]

%%%\fnmark[*]

\ead{katarina.doctor@nrl.navy.mil}

%\ead[url]{https://www.}

\credit{overall ideas, conceptualization, writing, editing, experimentation}
%and methodology of the study;  performed background research, contributed writing to all sections of the paper, performed calculations for case studies and led the overall writing effort.}

\affiliation[1]{organization={Naval Research Laboratory},
            addressline={Navy Center for Applied Research in Artificial Intelligence}, 
            city={Washington},
%          citysep={}, % Uncomment if no comma needed between city and postcode
            state={D.C.},
            postcode={20375}, 
            country={USA}}
%------- author 2 ------------
\author[2]{Mayank Kejriwal}[orcid=0000-0001-5988-8305]

% Footnote of the second author
%\fnmark[]

% Email id of the second author
\ead{kejriwal@isi.edu}


\credit{writing, editing}

\affiliation[2]{organization={University of Southern California},
          addressline={Information Sciences Institute}, 
            city={Los Angeles},
%          citysep={}, % Uncomment if no comma needed between city and postcode
          state={CA},
        postcode={90292}, 
           country={USA}}

%------- author 3 ------------
\author[3]{Lawrence Holder}[orcid=0000-0002-6586-3144]

% Footnote of the 3rd author
%\fnmark[]

% Email id of the 3rd author
\ead{holder@wsu.edu}

% URL of the 3rd author

\credit{writing, editing}

\affiliation[3]{organization={Washington State University},
           addressline={School of EECS}, 
            city={Pullman},
%          citysep={}, % Uncomment if no comma needed between city and postcode
          state={WA},
          postcode={99164}, 
            country={USA}}

%------- author 4 ------------
\author[4]{Eric Kildebeck}[orcid=0000-0001-7209-0226]

% Footnote of the 4th author
%\fnmark[]

% Email id of the 4th author

\ead{eric.kildebeck@utdallas.edu}

% URL of the 4th author

\credit{writing, editing, experimentation}

\affiliation[4]{organization={The University of Texas at Dallas},
           addressline={Center for Engineering Innovation, 800 W. Campbell Road}, 
            city={Richardson},
%          citysep={}, % Uncomment if no comma needed between city and postcode
          state={TX},
          postcode={75080}, 
            country={USA}}

%------- author 5 ------------
\author[1]{Emma Resmini} %[orcid=0000-0001-xxxx-xxxx]

% Footnote of the 5th author
%\fnmark[]

% Email id of the 5th author
\ead{emma.resmini@nrl.navy.mil}

% URL of the 5th author

\credit{experimentation}
%Performed background research, provided calculations for case studies;   contributed to the text of the paper.}

%------- author 6 ------------
\author[3]{Christopher Pereyda} %[orcid=0000-0001-xxxx-xxxx]

% Footnote of the 6th author
%\fnmark[]

% Email id of the 6th author
\ead{christopher.pereyda@wsu.edu}

% URL of the 6th author

\credit{writing, editing, experimentation}

%\affiliation[3]{organization={Washington State University},
%           addressline={School of EECS}, 
%            city={Pullman},
%          citysep={}, % Uncomment if no comma needed between city and postcode
%          state={WA},
 %         postcode={99164}, 
 %           country={USA}}

%------- author 7 ------------ 
\author[4]{Robert J. Steininger}[orcid=0000-0002-4316-9557]

% Footnote of the 5rd author
%\fnmark[]

% Email id of the 3rd author
\ead{robert.steininger@utdallas.edu}

% URL of the 3rd author

\credit{writing, editing}

%\affiliation[7]{organization={The University of Texas 
%at Dallas},
%          addressline={Center for Engineering Innovation, 800 W. Campbell Road}, 
%          citysep={}, % Uncomment if no comma needed between city and postcode
%          state={TX},
 %         postcode={75080}, 
  %          country={USA}}

%------- author 8 ------------ 
\author[5]{Daniel V. Olivença}[orcid=0000-0001-5474-2657]

% Footnote of the 5rd author
%\fnmark[]

% Email id of the 3rd author
\ead{dolivenca3@gatech.edu}

% URL of the 3rd author

\credit{writing, editing}

\affiliation[5]{organization={Georgia Institute of Technology},
           addressline={The Wallace H. Coulter Department of Biomedical Engineering}, 
            city={Atlanta},
%          citysep={}, % Uncomment if no comma needed between city and postcode
          state={GA},
          postcode={30332}, 
            country={USA}}

\cortext[1]{Corresponding author}



\begin{abstract}
Artificial Intelligence (AI) systems planned for deployment in open-world learning (OWL) environments and in real-world applications are frequently researched and developed in closed simulation environments, where all variables are controlled and known to the simulator or labeled benchmark datasets are used. Transitioning from these simulators, testbeds, and benchmark datasets to open-world domains poses significant challenges to AI systems, 
including significant change or increase in the complexity of the domain and the numerous novelties that are present in a real-world. The open-world  environment contains numerous out-of-distribution elements that are not part of the AI system's training set. 
Here, we propose a general, domain-independent estimation of domain complexity. Our first contribution is to systematically distinguish between two aspects of domain complexity: intrinsic and extrinsic. The intrinsic domain complexity is the complexity that exists by itself without any action or interaction from the OWL agent, an AI agent performing a task on that domain. This is an agent-independent aspect of the domain complexity. In contrast, the extrinsic domain complexity depends on the OWL agent and its skills. The combination of intrinsic and extrinsic elements captures the overall complexity of the domain. We frame the components that define and impact domain complexity levels in a domain-independent manner. 
We then describe three categories of measures (dimensionality, sparsity, and diversity) and argue that, taken together, they offer a comprehensive view of domain complexity. Last, we demonstrate these measures in several case studies in action, perception, and data science domains.
Domain-independent measures of complexity could enable quantitative predictions of the difficulty posed to AI systems when transitioning from one testbed or environment to another, avoiding bias when facing out-of-distribution data in open-world tasks, and when navigating the rapidly expanding solution and search spaces encountered in open-world domains.

% We organize these components into groups. We then describe methods for representing these components in a measurable form for estimating domain-complexity level. 
%We define intrinsic elements of a task within a domain that impact task difficulty and extrinsic elements of the AI system performing the task, which combined capture the overall complexity of the domain for a given task and agent. 

% Transdisciplinary domains' level of complexity differs in different ways. They simple, complicated or complex in different ways. We came up with a framework to define a common way estimating the level of complexity of domains. 
\end{abstract}


\begin{highlights}
\item A comprehensive introduction to understanding the complexity of domains in a systematic and domain-independent manner for the Artificial Intelligence community. 
\item A blueprint, and set of implementable measures, for systematically understanding domain complexity.
\item A systematic distinguishing between two aspects of domain complexity: intrinsic (agent-independent) and extrinsic (agent-dependent). 
\item Domain complexity measures spanning both action-based and classification-based domains.
\end{highlights}

% Keywords

% \maketitle
\maketitle
%== 1 ===================================================
\section{Introduction}
\label{1-Introduction}
When designing AI systems that can operate in open-world settings, it is important to be aware of the complexity level of the domain for which the AI system is built and the complexity level of the domain where it will be applied or transitioned. 
Understanding domain complexity helps in transitioning from theory, to simulations, to open-world and real-world domains. It helps to understand the boundaries and limitations, and it decreases uncertainties. It could prove indispensable to building an AI agent that can work in many domains, as well as for building agents that are able to operate in a single `open-world' domain where unexpected and novel events occur with non-trivial frequency. 

Self-driving cars, which have been developed and evaluated on closed courses or in simpler highway-driving contexts, will encounter new challenges while attempting to safely navigate open-world streets \cite{Claussmann2017}. Machine learning (ML) tools in medical contexts may encounter unexpected edge cases, noisy or distorted sensor data, and potentially significant differences in data distributions when transitioning from academic benchmark data to real-world use \cite{Holzinger2017}. 

In general, agents in simulated environments navigate a much smaller set of possible states and perform deliberative-reasoning search tasks over a much smaller set of possible state-action paths than what happens in the open world of a non-simulated, real environment. As one example of an AI agent transitioning from a closed-world domain to an open-world domain, \cite{Wilson2014} developed a motion and task-planning system for autonomous underwater vehicles (AUV) using a common simulation platform that they moved from development to deployment. If one takes the AUV agent, which has been built within a less complex simulated environment, installs it on a robot, and lowers it off the side of a ship into the real world, it can have significant problems correctly navigating in that real-world state space if not understanding its vastness, boundaries, and limitations. 

In the ever-changing open-world domains, there are a plethora of novelties that will have effects on an AI agent that has been trained in a closed-world setting. The overall level of complexity also changes when transitioning from a closed world to an open world. Transitioning from a closed world to an open world does not necessarily mean transitioning from a simple system to a complicated or complex system. However, understanding the boundaries of the domains an agent is facing will help with accommodating novelties in an open world. Knowing the complexity levels of both the domain from which an AI agent is transitioning from and the domain to which that agent is transitioning to will help avoid biasing the difficulty of detecting, characterizing, and accommodating novelty, thereby increasing the robustness of the OWL agents when performing a task in a domain of interest. Furthermore, understanding the complexity of the domain helps form anticipatory thinking in a multi-agent setting: anticipating the thinking about the actions, goals, plans, and strategies other agents might have. 

The core concept of complexity levels transcends the boundaries of a domain. In this paper, we define the components to consider to understand domain complexity and a framework for estimating the level of complexity of domains in an interdisciplinary, domain-independent light.

% e.g.  self driving car on road in a rural setting vs. busy street in major city 
% e.g. tic-tac-toe  vs.  monopoly
% e.g. 
\begin{figure}[htb]
  \centering
  \includegraphics[width=16 cm]{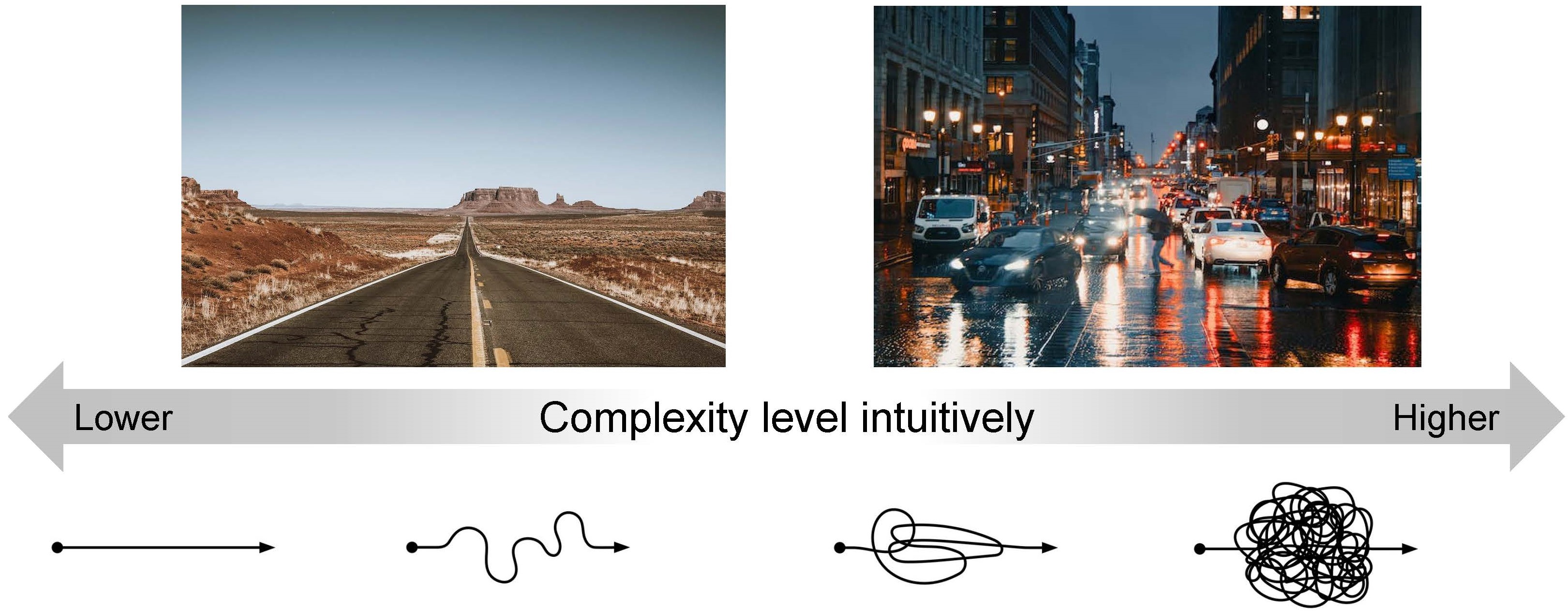}
  \caption{ We know intuitively that driving a car on a long road in a rural setting involves a lower complexity level than driving at night in a busy city where the likelihood of interacting with pedestrians crossing the road, other drivers, traffic lights, crosswalks, bicyclists, children playing, etc. is much higher. Similarly, getting from one point to the other in a straight line is intuitively less complicated than in a convoluted pattern. 
  }  \label{fig:intuitiveComplexity}
\end{figure}

Intuitively, we know or assume that complexity levels of domains differ. We know that tic-tac-toe is a simpler game than chess or Monopoly; classifying one dolphin in an image with a clear water background is less complicated than classifying all of the elements in a coral reef including dolphins and fish; and a self-driving car on a long straight road in a rural setting faces lower complexity level than driving on a street in a busy city (Figure \ref{fig:intuitiveComplexity}). 
We want to understand objectively and structurally the different aspects of domain complexity levels. Complexity levels of domains and tasks within domains can differ significantly when viewed from different perspectives. 
There are two parts of our contribution to the discipline: (1) components to consider for understanding domain complexity, and (2) estimating the complexity level of a domain. 
In this paper, we first frame the components that define domain complexity levels with a domain-independent and interdisciplinary perspective. We then describe the three measure categories 
% measure groups/categories
that together express the complexity level of a domain.  Last, we demonstrate these measures in several case studies in action, perception, and data science domains. 

%{\color{magenta}Larry noted: "This paragraph can be expanded to summarize the contributions of the work." }   

%The term "complex" is used many ways, even a salad dressing can be a complex one. First we will define what we mean by domain complexity level. .... 
% KD-TODO: finish the thoughts

%\paragraph{ } 

%== 2 ===================================================
\section{Motivation and Background} 
\label{2-Motivation and Background}

\paragraph{From Complexity to Difficulty} 
%"Complexity describes the thought process that the brain uses to deal with information. Difficulty on the other hand, refers to the amount of effort that the learner much expend within level of complexity to accomplish a learning objective." from David A. Sousa, 2006, "How the Brain Learns"
Understanding the complexity level of a domain helps assess and predict the difficulty of detecting, characterizing, and accommodating a novelty in that domain. Here we define a novelty as any element of the environment that was not present in previous training environments experienced by an OWL agent. Complexity and difficulty are different mental operations. The complexity level of a domain will affect the difficulty of detecting novelty, characterizing it, and accommodating it. Complexity also will influence the difficulty of generating novelty.
``Complexity'' is the description of a state --- the space of possibilities --- whereas ``difficulty'' relates to the challenge posed by a specific novelty within this space of possibilities. Estimating and understanding the complexity level of a domain is an important prerequisite for robust transitions between both domains and open-world learning environments. In contrast to estimating domain complexity, \cite{Ethayarajh} suggest a method for estimating dataset difficulty by establishing a framework with respect to a model \emph{v} that analyzes \emph{v-usable information} and \emph{pointwise v-information} to generate a quantitative value for defining dataset difficulty.

\paragraph{Bias and Inductive Bias}
A lack of understanding of the domain complexity in which the AI agent is performing a task can introduce bias when training OWL agents and when novelty is generated for building domain simulations \cite{leavy}. 
% bias is something to be avoided
Whereas understanding the domain complexity helps define inductive bias to be used by the AI agent for more robust performance.  
Understanding the complexity level of a domain one is applying an AI system to greatly decreases the bias of success toward specific data sets, while it also strengthens the inductive bias, as more information is understood to be able to make more helpful assumptions.

%Incorporating strong inductive bias into the OWL agent 
Inductive bias is described as factors that lead a learner to favor a path that are independent of the %observed 
data that have no yet been encountered \cite{GRIFFITHS2010357} and it is at the center of extending use cases beyond known data that is sampled from a domain \cite{Mitchell2007TheNF}. Understanding domain complexity enables building strong inductive bias and being able to create a generalization that is valid both for observed training data, as well as novelties that have not yet been encountered by the OWL agent \cite{Baxter}.

%Understanding the complexity level of a domain strengthens the inductive bias as it enables more helpful assumptions to be made. 
%The more that can be understood about the complexity of an AI system’s domain, the stronger the inductive bias of the learning algorithm will be, allowing for greater success when encountering novelties,

% cite these (Roman): these have been cited
% 4. Haussler, D. 1988. Quantifying inductive bias: AI learning algorithms and Valiant’s learning framework. https://www.sciencedirect.com/science/article/abs/pii/0004370288900021 
% 5. Leavy, S.; O’Sullivan, B.; and Siapera, E. 2020. Data, Power and Bias in Artificial Intelligence. https://arxiv.org/abs/2008.07341

%\paragraph{Domain complexity versus algorithmic complexity}  [few sentences, so one do not mix it up]
%Definitions of complexity used in theoretical computing are not precisely what is needed to understand complexity for real-world AI.  For one thing, they tend to compare complexity asymptotically, treating all constants as equivalent  \textcolor{blue}{[reference needed]}, where in AI searching $2^{10}$ states is a very different problem than searching $2^{1000}$. 
% CT do you have reference for this?
% The curse of dimensionality - should we describe this in short?

%== 2.1 =====================================================
\subsection{Why is a Systematic Understanding of Domain Complexity Important?}
\label{Benefits of Knowing Domain Complexity}

% Why is important to know the complexity of a domain? Why should we care? 
% (from science research perspective and from SAIL-ON or other program perspective)
% why complexity matters in terms of designing Artificial Intelligence (AI) for open-world domains?]
% importance of knowledge about the complexity of a system
% Problems that happen when you develop AI without knowing the complexity 
% estimating complexity level in a domain independent manner

% TODO: organize to 2-3 main points, highlight those and reorder to group around those main points

%[What goes wrong if you develop a solution without regards to the complexity of the domain? 
%Risks of not knowing the domain complexity]

%[Benefits of knowing / being aware of domain complexity:
%Robust agent for detecting, adapting to new situations
%Something related to resources, such as computational cost, GPU time?]
% CT: More generally-- e.g. Game AI, Self-driving Cars, explainable AI, medical AI, Robotics AI 
In the UAV example \cite{Wilson2014}, the simulated environment assumed that there was a 1-to-1 mapping between perception and the state space; i.e., that a change in perception was due to a meaningful change in state. Instead, in real-world contexts, with noise and complex factors related to environment, there may be many different perception values that all indicate effectively the same state for the goal reasoner. The real-world perception state space is effectively much larger than the space over which goal reasoning is actually performed. An AI system that was developed and tested in a simple, simulated perception space will tend to fail when it is put into the open sea, incessantly reporting discrepancies between its sensors and its model of the world, potentially getting trapped in a loop of constant replanning. As a result, the robot can freeze, drift off course, or lose contact with its users. This can be a very expensive problem.  
%     excerpt from the paper: "First, in stochastic or partially observable domains, precise modeling can lead to false discrepancies, which require extra computation and may be detrimental to agent performance. For instance, while executing a motion action, ocean currents may cause an AUV to move off its projected course. Reactive motion controllers can adjust the vehicle’s controls and correct this deviation without intervention from the GDA Controller. However, the Discrepancy Detector may incorrectly treat this deviation as a discrepancy requiring reevaluation of the agent’s goals. Employing a threshold during discrepancy detection is a common approach to such challenges, but a threshold value may not generalize to all continuous features in a domain." 
Addressing this issue required the development of new logic for dealing with the true complexity level of the perception space. Perception information was processed with bounding boxes to reduce its complexity sufficiently so that the existing goal-reasoning and planning logic could operate over it effectively. This measure allowed the robot to operate successfully in the open water. It is important to note that this issue was task-independent. There was effectively no nontrivial task that the robot could execute successfully before its algorithm had been modified to address the increase in domain complexity compared to the simulator domain. 

%To be cleaned up: 
Outside of the context of real-world robotics with deliberative reasoning \cite{Ingrand2013}, the domain complexity problem arises in other application areas as well. In domains with game states AI, the performance of Monte-Carlo Tree Search (MCTS) will depend on the size and complexity of the game tree. If the space of possible states and actions becomes excessively large, then the probabilistic exploration of the tree will have a higher probability of failing to sample the optimal paths, and then the agent may select moves that are suboptimal, poor, or even absurd. If the complexity of the game is understood correctly during AI development, then modifications can be made to improve the algorithm performance, such as using domain knowledge to bias search in games with large branching factors \cite{Chaslot2008}.

More dangerously, the problem arises in self-driving vehicles. \citeauthor{Claussmann2017} (2017) exhaustively surveys, categorizes, and evaluates diverse approaches to autonomous driving, but limited to only highway environments, and with respect to only eight simple maneuvers (such as changing lanes or exiting). This analysis does not cover important complexities in the highway domain, such as encountering an obstacle in the road or another vehicle merging into your lane without seeing you. Furthermore, it obviously will not apply to city street environments, which have far more states and complex transitions. Techniques that perform well in the simple, limited, highway domain may have very different properties and limitations than techniques that excel in a more complex environment, and failing to understand the impact of domain complexity on algorithm choice could lead to suboptimal decision-making and potentially fatal consequences.

The issue of correctly addressing real-world domain complexity also arises in data science contexts. In medical applications, explainable AI systems are increasingly being deployed to assist test analysis and decision-making \cite{Holzinger2017}. Systems that were designed for simple data distributions over one or two educational-benchmark datasets will fail or become extremely sensitive to data preparation and parameter tuning. As the feature set grows more complex, evaluation becomes more rigorous, or the data distribution becomes more diverse and heterogeneous. Again, systems developed on simple, toy-world domains, when moved to much larger, open-world domains without consideration for the change in complexity of the problem definition, will have failures with real-world consequences \cite{Holzinger2017}. Additional issues arise when considering the complexity of data itself, which is influenced by a variety of factors including dimensionality, scale, and mixed sources. The corresponding complexity level of the domain must be considered when working with this data for more accurate results. For example, some existing analytical methods rely on assumptions that do not apply to ``Big Data.'' Depending on the assumptions and domain of interest, the analysis could result in unusable information due to inaccuracy or ambiguity. Therefore, data science methods can benefit significantly from understanding domain complexity as it would provide broader insights for more optimal decision making, potentially in terms of business processes, methodological analysis, or problem solving in general \cite{Cao_2020}.

Similar to data science contexts, the social sciences can also benefit from understanding domain complexity in terms of computational text analysis, as its effectiveness can rely heavily on domain assumptions and computational and statistical model complexity. Better understanding these assumptions and levels of complexity would aid in choosing the best method for analysis, which would significantly increase analytical efficiency \cite{o2011computational}. Further evaluation of complexity can be extended to the field of systems engineering as well. Concretely evaluating the complexity of a system using standardized ``system complexity factors'' can greatly benefit the effectiveness of life cycle engineering and system management \cite{Potts_et_al}. The increasing prevalence of complexity studies in a wide variety of fields illustrates the usefulness of understanding complexity, especially in real-world applications.

\subsection{SAIL-ON} % maybe different section title would be better here

An application for quantifying domain complexity is the Science of Artificial Intelligence and Learning for Open-world Novelty (SAIL-ON) Defense Advanced Research Projects Agency (DARPA) program \cite{Senator2019}, which aims to research and develop the underlying scientific principles, general engineering techniques, and algorithms needed to create AI systems that act appropriately and effectively in novel situations that occur in open-world domains, which is a key characteristic needed for potential military applications of AI. The focus is on novelty that arises from violations of implicit or explicit assumptions in an agent’s model of the external world, including other agents, the environment, and their interactions. Specifically, the program will: (1) develop scientific principles to quantify and characterize novelty in open-world domains, (2) create AI systems that act appropriately and effectively in open-world domains, and (3) demonstrate and evaluate these systems in multiple domains.
The SAIL-ON program is divided into two groups: (a) those that facilitate the evaluations by providing novelty generators in the chosen domains across levels of the novelty hierarchy (see Table \ref{fig:Novelty Hierarchy table}), and (b) those that develop agents that can detect, characterize, and accommodate novelty. In each phase of the program, it is expected that the domain-independent characterization of novelty will be improved and that increasingly sophisticated and effective techniques for recognizing, characterizing, and responding to novel situations across domains will be developed. The novelties are categorized in a hierarchy representing the fundamental elements that make up an open-world domain, and by definition, every element within an open-world domain must have characteristic attributes and must be represented in some way. The open-world novelty hierarchy levels are: object, agent, actions, relations, interactions, rules, goals, and events (Table \ref{fig:Novelty Hierarchy table}). SAIL-ON performers have refined the novelty levels by adding difficulty levels: easy, medium, and hard. These difficulty levels refer to novelty accommodation. We are treating these hierarchy levels as part of the components to consider when estimating the complexity levels of domains, which we describe in Section 3 of this paper.

\setlength{\tabcolsep}{10pt} % Default value: 6pt
\renewcommand{\arraystretch}{1.5} % Default value: 1
\begin{table}[ht]
    \scriptsize
    \centering
    \begin{tabular}{c|c|p{0.4\linewidth}}
        \toprule
        \multicolumn{3}{c}{\textbf{Open-world  Novelty Hierarchy}}\\
        \midrule
        
        {} & 1 & {\textbf{Objects:} New classes, attributes, or representations of non-volitional entities.}\\ \hhline{|~|-|-|}
        \multirow{-1}{.15\linewidth}{Single Entities} & 2 & {\textbf{Agents:} New classes, attributes, or representations of volitional entities.}\\ \hhline{|~|-|-|}
        
        {} & 3 & {\textbf{Actions:} New classes, attributes, or representations of external agent behavior.}\\ \hhline{|-|-|-|}
        
        {} & 4 & {\textbf{Relations:} New classes, attributes, or representations of static properties of the relationships between multiple entities.}\\ \hhline{|~|-|-|}
        
        \multirow{-1}{.15\linewidth}{Multiple Entities} & 5 & {\textbf{Interactions:} New classes, attributes, or representations of dynamic properties of behaviors impacting multiple entities.}\\ \hhline{|-|-|-|}
        
        {} & 6 & {\textbf{Rules:} New classes, attributes, or representations of global constraints that impact all entities.}\\ \hhline{|~|-|-|}
        \multirow{-1}{.15\linewidth}{Complex Phenomena} & 7 & {\textbf{Goals:} New classes, attributes, or representations of external agent objectives.}\\  \hhline{|~|-|-|}
        
        {} & 8 & {\textbf{Events:} New classes, attributes, or representations of series of state changes that are not the direct result of volitional action by an external agent or the SAIL-ON agent.}\\
        
        \bottomrule
    \end{tabular}
    \caption{Open-world  novelty hierarchy levels developed by SAIL-ON Novelty Working Group. See acknowledgments for full list of contributors. }  \label{fig:Novelty Hierarchy table}
\end{table}

Complexity levels of domains are estimated in different ways for different domains, and it is challenging to generalize complexity estimation across multiple diverse domains. The structure of the SAIL-ON program highlights this need, as we would like to compare both domains and their complexity levels as well as to estimate the complexity level of generated novelties, which will enable us to predict the difficulty of detecting, characterizing, and accommodating novelty.

%\paragraph{Domain versus dataset} 
%Define what we mean by "domain" vs. "dataset" (Maybe: Data is a sample from task domain. Task domain is part of a domain with "goal" in mind. Dataset is a processed (probably labeled) data from a task domain.) 

\paragraph{Theoretical Frameworks for Open-World Novelty}
Some theoretical frameworks have been proposed for open-world novelty. \citeauthor{Langley_AAAI_2020} (2020) proposes a framework for characterizing open-world environments with goal-directed physical agents and how those environments can change over time. \citeauthor{Boult_AAAI_2021} (2021) propose a framework for defining theories of novelty across domains. This framework measures novelty based on dissimilarity measures in the world space and the observation space. These frameworks describe novelty in open-world learning from the perspective of an agent's performance, but do not offer an assessment of the complexity of the novelty's domain, which is the purpose of our paper. In this paper, we distinguish between agent-independent and -dependent parts of the domain and define the domain complexity components from both perspectives.
%These frameworks can be used to help define a measure of domain complexity in the context of an agent's performance, which differs from the approach here that defines complexity in an agent-agnostic framework.

%== 2.3 ===================================================
\subsection{Perspectives on Complexity from Different Disciplines}
\label{4-Perspectives on Complexity from Different Disciplines}
We briefly review existing approaches to understanding and measuring complexity as it applies to relevant computational disciplines: classical AI, data science, and systems research. These existing perspectives are important to take into consideration, but none provide the comprehensive, domain-independent analysis of the complexity level necessary to support the development and transition of AI to open-world domains.

\paragraph{Classical AI}
 \citeauthor{Hernandezorallo2010} (2010) use the term ``environment complexity'' to refer to increasingly complex classes of domain/task pairs. Their thesis suggests more intelligent agents are able to succeed in more complex environments. Their complexity level increases depending on the domain's number of possible states, the number of transitions (agent actions that change the state), and the difficulty of reaching the objective (in general, winning the game). The more possible states and the more difficult-to-select actions to navigate optimally through them, the more complex the problem. 

 Relating domain complexity to the state transition graph is a common approach in classical AI systems.  \citeauthor{Ingrand2013} (2013), \citeauthor{Chaslot2008} (2008), and \citeauthor{Claussmann2017} (2017) give examples of agents whose performance are, in fact, dependent on the environment complexity as used in \citeauthor{Hernandezorallo2010} (2010). These systems are sufficiently intelligent to solve problems of a certain complexity, but may fail as that complexity increases. 
 \citeauthor{Pereyda2020} (2020) propose a theory for measuring complexity by taking a resource-requirements approach, focusing on the three spaces: task, solution, and policy. The authors are relating complexity to the minimum description length of agents necessary to achieve different levels of performance on the task in a domain.
 
 While these approaches are relevant to domain complexity in our context, they focus on agent-dependent, task-specific complexity and do not explicitly address intrinsic domain properties that are independent of agent and may impact solution performance independently of the task.  A domain-independent complexity level needs to take into account a wider set of components in order to fully support development for open-world domains.

\paragraph{Combinatorial Game Theory}
The field of combinatorial game theory mainly focuses on sequential games, in which players alternate turns. A combinatorial game continues until there is no legal move or position left available for the player whose turn it is (\citeauthor{albert2007lessons}). Combinatorial game theory places an emphasis on descriptive theoretical results, such as measures of game complexity. Within game complexity, two measures can be utilized: state-space complexity and  game-tree complexity. The state-space complexity of a game is the number of legal positions from the starting position of the game. For some games, it is not feasible to calculate the number of legal positions, so instead an upper bound is computed with the understanding that this value contains illegal plays. The game-tree complexity of a game is the number of leaf nodes in the game tree from the initial position. To calculate the game-tree complexity, two additional values are needed: the branching factor of the game tree and the average game length (\citeauthor{Allis1994Searching}).

There are several ways that game complexity can be measured, each giving unique information about the game. Game tree size can be described as the total possible ways that the game can be played. This equates to the number of leaf nodes in the game tree that are rooted in the game's initial position. This can be an issue for some games that do not have a limited amount of moves. For games like this, the game tree size is generally assumed to be infinite. Decision trees are a sub-tree of the game tree and a way for describing the game winner, loser, or if the game is a draw. This is done by labeling positions in the game as such. Decision complexity and game-tree complexity are subsections of decision trees. Decision complexity is the number of leaf nodes within the smallest decision tree that also has the value of the starting position. Game-tree complexity describes the number of leaf nodes within the smallest full width decision tree that also has the value of the starting position. Game-tree complexity can be used for doing a minimax search to find the value of the starting position. Sometimes the game-tree complexity can be approximated as the game's average branching factor raised to the power of the number of plies in an average game. Computational complexity is a way to describe the asymptotic difficulty of a game when it grows arbitrarily large. However, this does not apply to all games, only to games that have been generalized to be made arbitrarily large.

\paragraph{Data Science}
 Data science focuses on fitting conceptual models to input data, allowing users to make predictions about new data points. Unlike classical AI, there is no agent that can take actions to interact with its environment; instead, the domain consists of a feature set and a distribution of data points across the feature space defined by that feature set. Complexity is determined by the difficulty of fitting meaningful, accurate models to these distributions so that they support correct predictions on new data. Properties such as the size of the feature space, the sparsity of the data, and the shape of the distribution impact the difficulty of this problem. 
\citeauthor{Remus2014} (2014) consider four task-independent domain complexity measures in the context of computational linguistics, focusing on the sparsity (word rarity), the feature set size (word richness), and the distribution of the input data (entropy and homogeneity). They find a strong correlation between domain complexity and the performance of a standard ML classifier on the data. Similar to Hernández-Orallo's observations on classical AI, as complexity increases, performance decreases. These general ideas are broadly applicable as sources of complexity in problem solving, as we will discuss in Section 6. However, by themselves, they do not constitute a general definition for domain complexity levels.

%Remus, R., \& Ziegelmayer, D. (2014). Learning from domain complexity. Proceedings of the 9th International Conference on Language Resources and Evaluation, LREC 2014, 2, 2021–2028. http://www.lrec-conf.org/proceedings/lrec2014/pdf/480\_Paper.pdf 480\_Paper.pdf (lrec-conf.org)
% is this domain complexity? if not, state what is it, how is differ from domain complexity

\paragraph{Complexity from cognitive science perspective}
%[Reference pandemic paper, using ontology from cognitive sciences, etc.]

In cognitive science, complexity is considered with analogy to human or animal intelligence.  Intuitively complex problems may take significant effort to solve correctly. \cite{Jackson2020} expands this idea to an ontology of definitions. \citeauthor{Legg2007} (2007) begins with analogies to various grades of animal and human intelligence, and draws on this to define a platform for evaluating machine intelligence for classical AI.  
% They note common traits such as [ add.... ].   [summarizing] considers arguably the opposite problem, explaining complex agent strategies to humans when the agents are operating with internal domain definitions that are larger or more complicated than humans can easily understand.   Again, factors such as [ add... ] impact the complexity of the AI system and the difficulty of developing a human-accessible summary. 

%Amir, O., Doshi-Velez, F., \& Sarne, D. (2019). Summarizing agent strategies. Autonomous Agents and Multi-Agent Systems, 33(5), 628–644. 
% https://www.researchgate.net/publication/334709443\_Summarizing\_agent\_strategies

% cite the pandemic paper on complexity here

\paragraph{Systems Research}
 Systems research considers complexity in the form of challenges that arise when organizing multiple interacting components, whether those components are team members, organizations, industrial production systems, software modules, or even elements of programming languages as interpreted by a compiler. In these contexts, patterns of dependencies between components are a key factor in the complexity of the problem. Large webs of interdependencies require significantly more computational time, or human cognitive load, to consider fully and to develop optimal solutions. If the interdependencies can be arbitrarily complex, then in general, the problem may be computationally hard (e.g., the knapsack problem). However, most real-world problem instances are tractable. 
 
 Various tools have been developed to help visualize, measure, and address complexity in system-design contexts. \cite{Stuikys2009} introduce three measures dealing with the cognitive and structural complexity of feature diagrams in order to estimate total complexity. \cite{Jung2020} propose a new metric based on the relationship between the requirements and elements of a system design. A series of matrices are then used to quantify the complexity of the system design. \cite{Prnjat2001} analyze the use of established software metrics to identify complexity and risk indicators early in the system development lifecycle. \cite{Kim2014} discusses the Cynefin Framework, which helps classify complex environments more specifically. 
 
 This work relates to our problem of domain complexity, specifically in the case of domains that contain multicomponent systems. For example, when an agent must navigate a multiagent environment with a large number of external entities, it can cause a combinatorial explosion in the set of possible states and transitions necessary to capture the interactions of those entities, resulting in a dramatically larger domain. This can be an important facet of domain complexity; however, there are many other sources of complexity that do not originate from interactions with multicomponent systems. Additional components are necessary to capture the complexity of scenarios, such as single agents navigating inclement environments, ambiguous perception data, or challenging game objectives. System complexity metrics, by themselves, do not constitute a domain-independent complexity level.

\paragraph{}
In this paper, we frame the components that define and impact domain complexity levels in a domain-independent light. We organize these components into groups. We then describe methods for representing these components in a measurable form for estimating domain complexity levels. 
%The result is a six-parameter variable chart of domain-complexity level for the six groups of components that we define in Section 5.

%== 3 ===================================================
\section{Framework for the Components that Define a Domain's Complexity Level}
\label{sec:Components}
\begin{figure}[htb]
  \centering
  \includegraphics[width=15 cm]{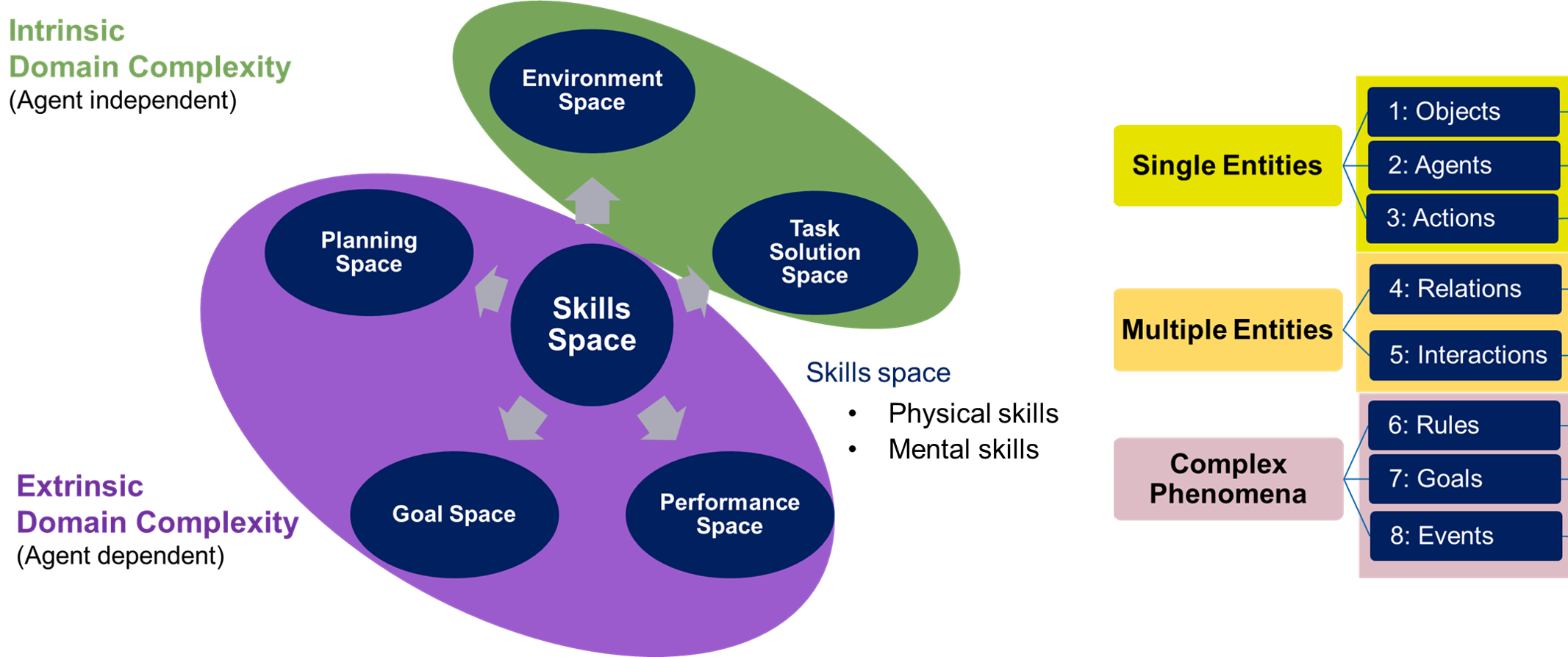}
  \caption{ Components that define domain complexity are extrinsic, depending on the mental and physical skills of an OWL-agent, and intrinsic, independent of these skills. Components can then be characterized further in ``environment,'' ``task solution,'' ``planning,'' ``goal,'' ``performance,'' and ``skills'' spaces. These ``spaces'' can be further divided into single- and multiple- entities and complex phenomena.
  }  \label{fig:components}
\end{figure}

There are two parts when determining the complexity of a domain: intrinsic and extrinsic. The intrinsic complexity is the complexity that exists by itself without any action or interaction from an OWL agent, which is the AI agent performing a task on that domain. It is an OWL-agent-independent aspect of the domain complexity. The intrinsic components that define and impact the complexity level of a domain are grouped further into ``environment space'' and ``task solution space.'' The extrinsic complexity of a domain is dependent on the established mental and physical skills of the OWL agent. It contains the ``performance space,'' the ``goal space,'' the ``planning space,'' and the ``skills space'' (Figure \ref{fig:components}).

Considering the complexity level only from an intrinsic, OWL-agent-independent perspective, or only from an extrinsic, agent-dependent perspective, in which we only observe the complexity level of an OWL agent's actions and existing knowledge, would result in a skewed metric of the complexity level. Therefore, we need to look at both the intrinsic and extrinsic parts of the domain complexity in order to get a balanced observation. % or result  
% Only intrinsic perspective will result too high dimensionality, e.g. looking all the pixels, and channels and parameters in imagery can result in very high numbers. 

\begin{figure}[htb] 
  \centering
  \includegraphics[width=5cm]{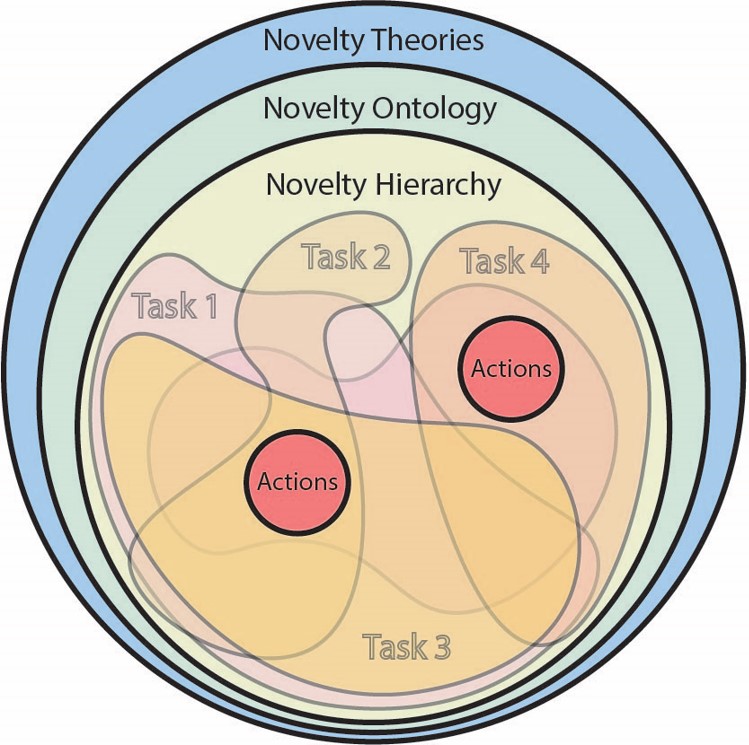}
  \caption{ Venn diagram for open-world novelty. Individual task domains can included hierarchy levels in different ways, such as how actions in Cart-Pole may include the player moving the cart where in Minecraft they may include enemy agents firing weapons. A given open-world task will include a subset of novelties within the space of possibilities defined by a hierarchy. Different tasks can overlap and intersect each other in terms of the types of novelties they are capable of including and how they can represent those novelties.}  \label{fig:Venn-OpenWorldNovelty}
\end{figure}

We took into consideration the open-world novelty hierarchy levels described earlier in the ``Motivation'' section as part of the components to consider when estimating complexity level. These hierarchy levels are: objects, agents, actions, relations, interactions, rules, goals, and events (Table \ref{fig:Novelty Hierarchy table}). We differentiate between novelty theories, ontologies, and hierarchy categories: theories can describe anything that is conceptually possible, ontologies classify elements that are realistic, and hierarchy categories represent a subset of realistic elements that are practically important and scientifically useful for open-world domains. Different tasks in different domains may or may not be affected by all the hierarchy levels in the same way, as shown in Figure \ref{fig:Venn-OpenWorldNovelty}.

We list these components in groups that define the intrinsic and extrinsic domain complexity levels. Each of these components has task-dependent and task-independent parts that we do not list separately here. 

Note that all of these components will not be present for every domain. Also, some tasks will not interact with some of these components even if they exist in the domain.

% Maybe we should mention that: It is not the same as the terms in strategy domains related to game theory: Endogeneous and Exogeneous complexity [in Chatterjee and Sabourian, 2008]. The Endogeneous and Exoneneous complexities are rather the OWL agent's constraints, not domain complexities.
% Endogeneous: the agent will choose the number of states
% Exogenous: constraining the agent with number of states, which will change the motivation for what strategy to use

% intro that we want to set the framework for components that define intrinsic and extrinsic domain complexity
% we will than use this framework of components and apply density estimation and diversity (homogeneity)  test from Information Theory and combine it with state transition graph. 

% TODO: 
% Delete redundancies in the groups
% review if the components are in the correct groups, are the examples accurate?
% go through each of the space and add example from 
%   - action/strategy domain
%   - perception / data science domain
%   - AUV from the anecdote 

\subsection{Components that Define Intrinsic Domain Complexity}
\label{subsec:Components that Define Intrinsic Domain Complexity}

The intrinsic, agent-independent complexity is structural and can have many relevant levels of description for computer simulations and for the natural world. There are task-independent and task-dependent parts of the intrinsic domain complexity. The task-independent part defines the broad structure of the domain complexity, whereas the task-dependent part is based on which aspects of the available complexity in the domain are relevant to, or utilized in, a given task. The elements contributing to intrinsic domain complexity are divided into the ``environment space,'' which includes all elements of the environment, and the ``task solution space,'' which includes only those elements relevant to completing a given task within that environment.

\subsubsection{Environment Space}
The environment may comprise parameters/variables, data schema, tokens/observations, scale size, objects, states, or agents internal to the system. The categories of the novelty hierarchy (Table \ref{fig:Novelty Hierarchy table}) represent the elements of an open-world environment space. The environment also comprises discrete entities (objects and agents), which can have static relationships with each other (relations) and dynamic interactions (interactions) that, in turn, can combine to create complex phenomena based on multiple relations and interactions, such as events. The complexity of the environment increases as the number of elements in each category, and the number of distinct attributes and representations for each element, increase.

\textbf{Single entities:}
The fundamental elements for an environment are the single, discrete entities present in the space, such as \textbf{objects}, \textbf{agents}, and \textbf{actions}. In action domains, this would be the number of unique objects and agents in a task environment, (e.g., blocks, tools, and enemy units in Minecraft). In perception domains, this would be the number of classes (supervised learning) or the number of clusters (unsupervised learning). Each of these categories can be expanded to include more discrete classes, and instances of each class can have an increasing number of attributes and representations. In the AUV example, an underwater simulator may have only one type of fish, with a single color and a single, swim-forward action that must be avoided by a submersible. A more complex simulator may have hundreds of types of marine life, each performing multiple actions and having variable visual appearances. Note that these agents are internal to the domain's system and are not the OWL agent, who is performing a task on the domain. Examples of agents that are internal in the domain are other players of a game, other drones in a swarm, or other sensors in the domain. Sensors that are internal in the domain would be, for example, a radar sensor that detects the OWL agent's presence or movement. 

\textbf{Multiple entities:}
When multiple entities are present, new elements of the domain emerge, including static \textbf{relationships} and dynamic \textbf{interactions}. In the AUV example, in a less complex simulator, all fish may be the same size (relationship) and may swim in a straight line (interaction). In a more complex simulator, marine life may include very small and very large animals, and fish may swim in complex schooling and swarming patterns. These elements are, by definition, only present and observable when multiple entities are present in the environment. Similar to the single-entity categories, the relationship and interaction categories can have a growing number of discrete classes, attributes, and representations as domain complexity increases. For instance, new interactions can be added, such as animals following each other, eating each other, fighting with each other, etc. The attributes of existing relationships and interactions also can increase in complexity, such as the speed and intricacy of schooling behavior.

\textbf{Complex phenomena:}
When multiple relations and interactions are present, complex phenomena can emerge within environments. Three examples of such complex phenomena are events, goals, and rules. 
\textbf{Events:}  When multiple interactions occur in series based on entity relationships, events can take place (e.g., a fire spreading across trees based on their physical proximity relationships). 
\textbf{Goals:}  the goals of agents in the environment also become discernible through the sequence of interactions the agent pursues within that environment. 
\textbf{Rules:} In real-world domains, and extensively in games, rules can be applied globally to define which relationships and interactions are permissible in the environment. Rules and constraints in the environment space reduce the size of the environment. These include the rules of a game, the number of states constrained by symmetry, the number of possible agent interactions, and the set of state transitions that are constrained by rules.

The following are some examples of rules: 
The rules of driving restrict which lanes can be used when driving in certain directions, the rules of Monopoly constrain spending by the amount of money the player has (0 money = 0 spending), and the rules of tic-tac-toe prevent a player from placing an X on a spot that already has an O. In AI simulations, real-world interactions or complex phenomena (e.g., a submersible requiring fuel to generate power) may be simulated as domain rules (e.g., an agent cannot issue move commands if its fuel level = 0).

For each category of elements that emerges with single entities, multiple entities, and complex phenomena, the number of distinct classes, distinct attributes, and distinct representations can be increased. Collectively, the scale and diversity of these elements determine the environment space. 
% The set of of all possible paths through the state transition graph. 

%The set of possible states
%The axis of the feature space
%Also constrained by rules
%\textbf{Constraints in the environment: }
%Constraints in the environment space reduce the size of the environment.
 %These are the number of states constrained by symmetry, number of possible agent interactions,  the set of state transitions that are constrained by rules, and relations. In perception domains this would also include the set of data classification classes.  Here would fall also the rules that define a game or the movement of a robot, or the rules of the traffic on a highway.

%These constraints declare what is nuisance (e.g. absurd, ridiculous, cannot happen) in the domain. 
% Examples: 
% the ocean constrains the speed of the agent, so it cannot move extremely fast; 
% the rules of monopoly constrain spending by the amount of money the player has (0 money = 0 spending);
% the rules of tic-tac-toe prevent a player from placing an X on a spot that already has an O;
% The constraints of reality mean that there cannot be a real-world image of a flying school bus. %

\subsubsection{Task Solution Space} 
The task solution space comprises the number and diversity of paths that can be taken to complete a task, whether in a real, open-world domain or in a simulator, where the available world states and action space are defined. This space increases in complexity as the set of possible state transitions increases and as the available paths for success become more complex (see example state transition graphs in Figure~\ref{fig:TTT_states}). In perception domains, the task solution space would also include the set of data-classification classes. The complexity of the task solution space is not dependent on the complexity of the environment, and the complexity of the environment may increase, decrease, or not impact the task solution space. For example, consider the domains of chess and a self-driving car in the real world. The environment of chess is quite limited, with a small number of unique objects and interactions, whereas the real world of the car may have thousands of unique objects, external agents, relationships, and interactions. If the task for the self-driving car is to move forward 1 meter, then this expansive environmental complexity has a minimal impact on the task. Furthermore, if the available action space includes a command to ``move forward 1 meter,'' then this task becomes trivial. Winning a game of chess against a challenging opponent, by comparison, would require much more computation and strategy, and would have a lower success percentage than the self-driving car's task. 

The number of possible paths, the set of possible agent interactions, and the restrictions on successful paths to achieve a goal are the primary drivers of complexity in the task solution space. We can consider a maze task in a grid environment: if the number of available paths increases from 2 to 10, then the complexity of the task solution space will increase if there is only one correct path, but will not become more challenging if every available path leads to the goal. Further, the addition of a new object class, such as a boulder, may increase the complexity of the maze environment while decreasing the complexity of the task solution space by blocking off incorrect paths and reducing the search space. 

The task solution space defines the number of possible actions and the distribution of paths (number of paths, number of intersections) through the state transition graph, the number and degree of dependencies and connections between agents and state transitions, and the degree of available strategies (defined as the set of all possible decision sequences through the environment, without violating any constraints). This set of paths can be represented as a state transition graph as explored in Section 4 ``Domain Representation.''

% Diversity
% measure for this would be entropy

%\subsubsection{Connectivity / Interactions space} % or connectivity?
%Connectivity and interactions that influence each other.
%Number of possible agent interactions, of possible actions, the set of state transitions
% number of possible connections / networks, of states, of moves
%The set of paths through the state transition graph that are depending on others. 
% Would this be related to upper bound in a lattice of the state transition graph?

%\subsection{Strategy / Interdependence} % Strategy? Policy? 
%Strategy is the set of all possible decision sequences through the environment without violating any constraints. 

%States that that aren't reachable in the the state-transition graph, and parts of the features space that are known to be empty. 
% Strategy, agent, policy is a model
% API - Application Programming Interface?

\subsection{Components that Define Extrinsic Domain Complexity}
\label{5.2-Components that Define Extrinsic Domain Complexity}
The components of extrinsic domain complexity are defined by the  OWL agent that is performing a task on this domain. The extrinsic domain complexity depends on the skills of the OWL agent and the skills needed to perform the tasks. This agent-dependent, extrinsic complexity is a physical and mental model of the OWL agent that has structural, observational, and planning components and is a subset of full domain complexity (Figure \ref{fig:components}). 

\subsubsection{Performance Space} %task scoring function
Performance space is the agent-policy scoring function and the performance of the OWL agent acting on the domain. These are the scoring of games, such as win/loss, win/loss/draw, or the range of score numbers. This would also be the reward function in reinforcement learning. 
For example, the performance space of an AUV could be the time taken to travel from point A to point B. 
 
%The set of all possible functions from a game trace or decision sequence to final score.   
%Sparsity and Heterogeniety applies here, and they talk about the difference between the 
% Dimensionality: The length of the successful paths through the state-transition graph or the size of the successful model in data science.  

\subsubsection{Goal Space} 
The goal space is the number of elements or the size of the possible goals in the domain. Some domains have only one goal, e.g., to win, to destroy, or to move from point A to point B. Other domains can have multiple goals, e.g., to win with the highest amount of money, to move from point A to B while having x amount of fuel left, or to avoid certain actions (e.g., car: crashing, toll roads, flooded roads; aircraft: storm cloud or all clouds in VFR flight conditions).
The goal space includes the set of possible strategies, the number of goal states, and the set of all possible paths through the state transition graph that end in a goal state. Not all domains have a goal space.

\subsubsection{Planning Space}
Planning space is the set of possible plans that can be generated. Planners are traditionally given as input: (1) a goal, (2) a set of
actions, and (3) an initial state. Many plans might accomplish the same goal, while a single plan might accomplish multiple goals. The size of the set of possible plans defines the planning space. Not all domains have a planning space. 

\subsubsection{Skills Space}
These are the components that define the skills that an OWL agent would need to perform a task in a domain of interest. They consist of physical and mental skills. 

\textbf{Physical skills (hardware):}
These are the physical abilities of the OWL agent --- the hardware. If a rock is in the moving agent's path, it would be insignificant if the agent is a bulldozer robot, but it would be significant if the agent is a vacuum-cleaner robot. Other characteristics include GPU, CPU, ROM/RAM, controller, type, number, accuracy, and precision of sensors or activators. The AUV has sensors, such as a camera, a pressure sensor, a sound sensor, etc. As another example, camera A might have only three (RGB) channels with a specific resolution, whereas another camera, camera B might have hundreds of channels with courser resolution.

\textbf{Mental skills (software):} % this needs review
These are the software or knowledge, cognitive skills, and intuitive skills that an AI agent needs for performing the task in the domain of interest. 
% adaptability (?),
This would include the ability to learn, predict, plan, set up, or follow goals. Other characteristics include the extension and complexity levels of interactions, 
%"ambiguity" might be a measure of difficulty for goals/classes
actions, perception, goal states, and events.

% NOTE: In addition to these components it would be valuable to incorporate in the future the related 
% Uncertainty
% Noise
% Partial data (e.g. partially observable image, when we do not have access to everything)
% Randomness
\paragraph{}
In summary, the described intrinsic and extrinsic components of the domain with the listed ``spaces'' define the complexity level of the domain. 
These components and the listed ``spaces'' will each be represented and estimated using the methods described in later Sections. The domain complexity level is expressed with a value for each ``space'' described above. Some of the ``spaces'' will not be applicable for some domains. Note that these components will have additional dimensions when we are considering a dynamic domain in which the complexity level becomes nonstationary. As the complexity level increases, the time scale will become more significant.

%== 6 ===================================================
\section{Domain Representation} 
\label{6-Domain Representation} 

%------------------------------------------------
%We are proposing general methods that can capture intrinsic and extrinsic domain complexity across variety of application contexts and disciplines.  [add: Argument that they're mutually exclusive (3D graph).]  These to the above components

The data science and perception domains versus the action and planning domains require different representation methods. The action and planning domains are often represented in form of state transition graphs. Whereas, for the perception and data science domains, the feature space provides a geometric representation for analyzing.
 
In this section, we briefly review the two fundamental domain representations for problems in action and planning domains and in data science and perception domains. % We then discuss several key factors of domain complexity that can be understood to hold over both of these problem contexts and domain representations. 

\subsection{State Transition Space} 
A state transition graph provides a conceptual representation of the domain, including possible world states, possible actions or transitions in each state, and the consequences of those actions on the state. In addition, the graph can represent tasks within the domain by indicating the initial state, the set of states that satisfy the current task objective (goal states), and the possible paths that lead from the initial state to the goal state. State transitions may occur as the result of the agent's own actions, opponents' or allies' actions, or external environmental actions. 

\begin{figure}[htb]
  \centering
  \includegraphics[width=16cm]{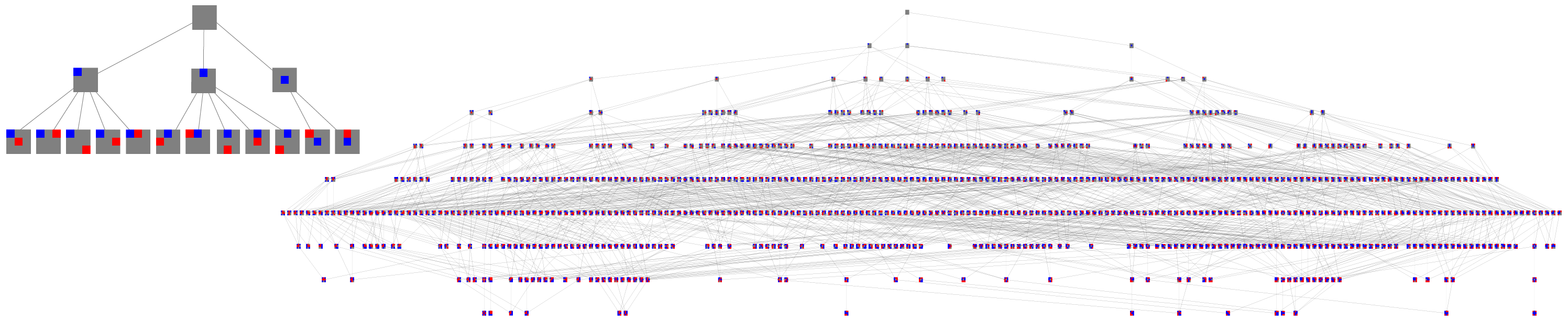} %{figures/ttt_game_state.png}%{figures/TTT_state_transitions.jpg}
  \caption{Visualizaiton of the state transitions for the tic-tac-toe game unique moves as decision tree. Each layer in the tree represents a move on the 3x3 board of the tic-tac-toe game and each gray square is a state/position on the game's board. The red and blue on the gray board shows the position of the two players. The left tree shows enlarged the starting and the first two states, that are the first two unique moves in the game. The right graph shows all the unique states of the tic-tac-toe game. The sequence of states formed by these possible moves is referred to as a search tree. Tracing a single path from the top of the search tree to the last layer or until a game is won or there is a tie, provides one possible outcome of a game. We can see after the initial position nine layers in the right figure, since the tic-tac-toe board has nine states/positions. The last layers have far fewer dots/positions because the game can be won without filling all the positions on the board.}

  %Visualization of the state transitions for the tic-tac-toe game as concentric circles (\cite{tictactoetransitions}). Each concentric circle represents a move on the 3x3 board of the tic-tac-toe game and each dot is a state/position on the game's board. The left graph shows two state transitions, the center is the empty board. The middle graph shows four moves on the board and the right graph shows all the state transitions. The number of state transitions quickly increases, from 1 to 9 to 72 and so on, until a game is won or there is a tie. The sequence of states formed by these possible moves is referred to as a search tree. Tracing a single path from the root of the search tree, that is from the center of the concentric circles to the most outer circle, provides one possible outcome of a game. We can see nine concentric circles in the right figure, since the tic-tac-toe board has nine states/positions. The outer-most circles have far fewer dots because the game can be won without filling all the positions on the board. } 
  \label{fig:TTT_states}
\end{figure}

State transitions are often illustrates as decision trees. Figure \ref{fig:TTT_states} shows the unique states in a state transition graph for the tic-tac-toe game. The starting point of an empty 3x3 board is on the top of the tree. Although the first move is anywhere on the nine spaces on the 3x3 board, the unique choices are only three, because the rest board positions are the transformations or symmetries of these unique three. The first move in the game therefore, gives three unique choices on the board, and the next move gives additional 12 unique options, as the left graph shows in Figure \ref{fig:TTT_states}. Tracing a single path from the top of the state transition tree to the last layer (ply) provides one possible outcome of a game. 

%State transitions are often illustrated as decision trees. Figure \ref{fig:TTT_states} depicts a state transition graph for the tic-tac-toe game as concentric circles. The starting point of an empty 3x3 board is in the center of a circle. The first move in the game gives nine choices on the board, and the next move gives additional 72 options, as the left graph shows in Figure \ref{fig:TTT_states}. The middle graph shows the first four moves, and the right graph shows all the possible states of tic-tac-toe, where each concentric circle is a move, or in other words, the state transition in the game. Tracing a single path from the center of the concentric circles to the most outer circle provides one possible outcome of a game.  

% A simplistic game like tic-tac-toe has approximately 6000 state transitions, but even its 3-dimensional version, Qubic, has $10^{30}$ state transitions.

This method of representing the domain allows us to characterize an AI action-and-planning task as identifying a potential path through the state space to the goal, successfully taking actions at each decision point to remain on an efficient, low-cost path that leads to the goal. This becomes more challenging as the complexity of the state transition graph increases; more states, more possible actions, very heterogeneous states and transitions, and sparser or more inaccessible goal paths all increase the difficulty of recognizing and maintaining an efficient, successful path to a goal state, which is to create a horizontal, vertical, or diagonal row of a player's mark in the case of tic-tac-toe.

% Elements in state transitions that matter:
% - tree width of the state transition graph (number of connections, edges and number of states:) 
% - percentage of dead-end states 
% sparsity of the paths to the goal
%  percentage of paths that lead to the goal 
%  percentage of states that are possible to succeed
% - width of the tree
% least c... ancestor
%   Do all possible edges matter or just those that are part of all possible paths? If later, than this would be task dependent. Wouldn't it? 
% - frequency of connections? 
% - do the upper and lower boundaries matter (as of in lattices)? (not really) 

\subsection{Feature Space}
A feature space is an abstract space represented as an $n$-dimensional matrix of all possible values for a set of features that characterize the data of a task in a domain, where $n$ is the number of features, or parameters, that each represent an axis of the space. The spatial representation of data from perception and data science domains enables introducing the statistical ``distance'' and ``closeness'' that expresses similarity based on features.   

\begin{figure}[htb]
  \centering
  \includegraphics[width=10cm]{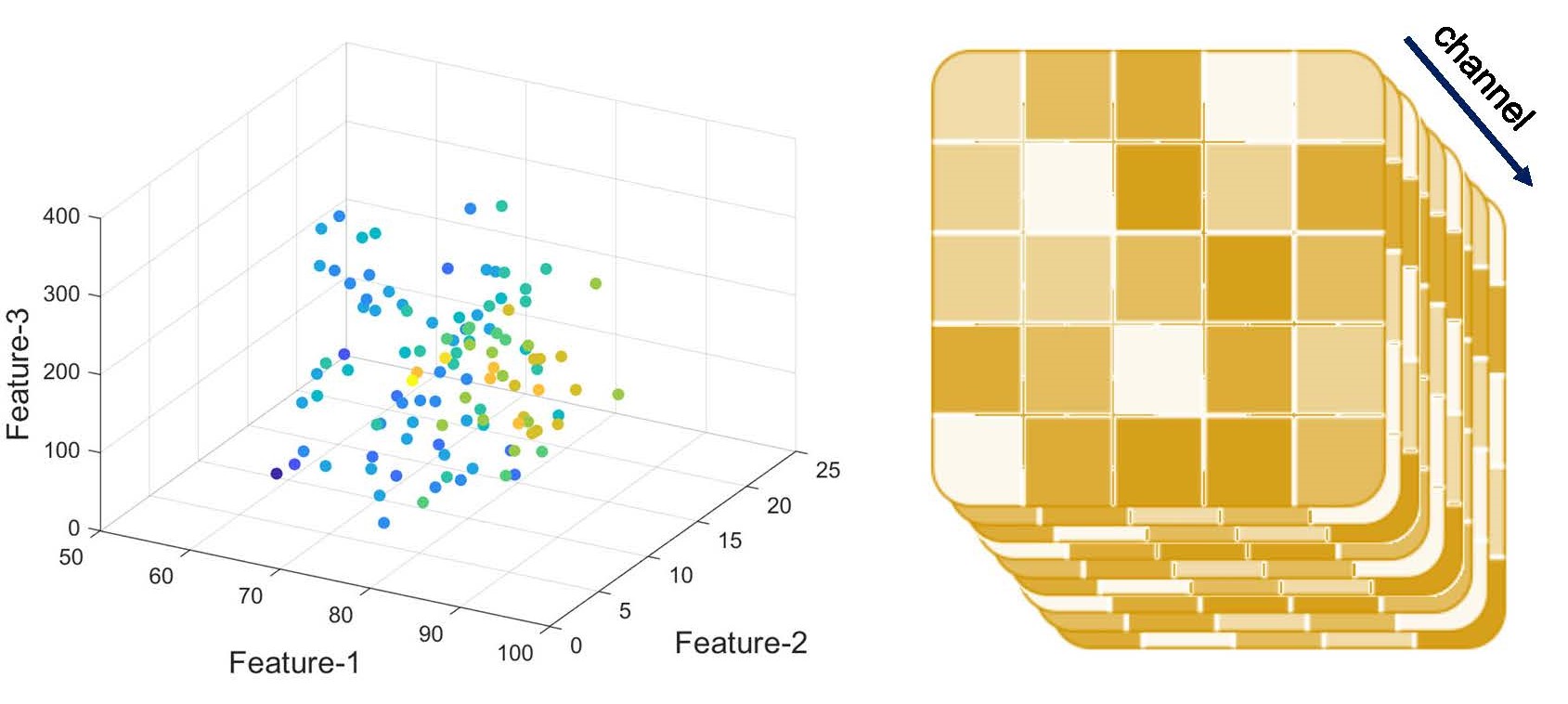}
  \caption{Left: An example of a 3-dimensional feature space for a dataset containing three features/parameters. Each observation in the training dataset can be plotted as a point in this 3-dimensional space. Machine Learning algorithms then can characterize the distribution of points across this space (using clustering, regression, classification, or other techniques) so that they are able to make predictions about how new data points might fit in the existing distribution. 
  Right: a feature space example from a perception domain. The figure shows an image of 5x5 pixels and n-channels. 
  } \label{fig:FSpaceExample}
\end{figure}

Figure \ref{fig:FSpaceExample} on left depicts a small, 3-dimensional feature space for a simple data science or perception domain. As the number of features in a data science problem gets larger, the dimension of the feature space grows. Higher-dimensional tasks are computationally more difficult for common model-fitting techniques, such as clustering and regression. Also, as a given set of data is spread over a larger feature space, it generally grows more sparse and the distribution can become harder to characterize accurately. 

Additionally, if the data is distributed more diversely across this space, then it can be difficult to characterize accurately; more diverse, outlying, or smaller subgroups of observations may not be classified correctly. Similarly to the inherent domain complexity due to the distribution of paths through the state transition graph, these are inherent difficulties with the data domain (feature space and training data distribution) that will have some impact on any classification or prediction task in that domain.

%== 5 ===================================================
\section{Complexity Measures}
\label{Complexity Measures}

Complexity measures compress information; they transform sets of entities into a single number. During this process, information is condensed, and some information is subsequently lost. 
Therefore, we should not expect a one size fits all measure. Instead, we will consider several complexity measures. The wide variety of measures will allow us to pick the most appropriate measure for a specific context, as well as offering a universal view of the domain. We will describe the estimation of the complexity level of a domain with a combination of three measures: dimensionality, sparsity, and diversity. 
It is more accurate to say ``estimate complexity level'' than ``measure,'' since we cannot directly measure, but rather, we estimate a complexity level of a domain. 

%Directly measuring the complexity level of a system isn't always possible. In most cases when investigating a system the complexity level can be estimated using a few different measures and equations, this doesn't give a clear answer as to which problem space is more complex every time though. The complexity level of domains can be expressed with the combination of three different measures: dimensionality, sparsity, and heterogeneity. Using each of these measures they can define the space that the agent resides to solve tasks in the domain. 

\subsection{Dimensionality}
In both classical AI and data science applications, dimensionality impacts the difficulty of completing tasks in the domain and is one of the key measurements for estimating domain complexity.  For intrinsic domain complexity (agent-independent) estimation we want to find the answer to the question ``How big is the problem space?''.  For extrinsic (agent-dependent) the question we want to answer is ``How long/large is the decision sequence needed to arrive at success?''. 

In classical AI, increasing the number of possible states (including increasing the environment size, the environment features, and the number of possible interactions with objects or external entities) increases the size of the state transition graph. Similarly, increasing the number of possible actions increases the breadth, or tree width, of the state transition graph. Both increase the dimensionality of the environment space and significantly increase the complexity and difficulty of tasks that involve navigating the state transition graph to reach a goal state. If a given goal state is accessible only by very long paths (which require the agent to make a long series of correct decisions), then this increases the dimensionality of the task solution and performance spaces, the complexity of the domain, and the difficulty of reaching the goal. 
In action spaces, the dimensionality can be described as the possible number of ``moves,'' or positions, agent(s) can make within the space. For example, in the field of game complexity, the calculation of state transitions is used to provide the dimensionality of a game.

In data science, increasing the number of possible features or the number of possible values for each feature (for example, by increasing the set of possible objects or environment conditions) increases the size of the feature space and the domain complexity with respect to the environment. This increases the difficulty of fitting models to characterize the distribution of the training data across that space. As vocabularies increase, sensor resolution increases, or additional variables are added, the dimensionality of the data science problem increases, and the task difficulty increases as well. Meanwhile, as the number of possible classes or prediction values increases, the set of possible parameterizations for fitting the model to the feature space can increase exponentially, increasing the complexity with regard to task solution and performance.   

Measuring game complexity is often described in two ways: state-space complexity and game-tree complexity. (\cite{enwiki:1129116028}) We estimated the dimensionality for several case study domains described in Section \ref{Case studies} using these measures. Because the calculation results are usually very large numbers, we are using the ceiling of the logarithm to base 10 of each, so the results are easier to compare. 

To calculate the state-space complexity, we need to know the board size, which is the number of unique positions on the game board. For example, a chess board has a board size of 64. The upper bound dimensionality is the environment space (Figure \ref{fig:components}), and the state space complexity is the task solutions space  of the domain described in Section 3.1.2. The upper bound is typically larger than the actual task solution space because it is not constrained by the task-related rules. When the rules are taken into consideration, the solution space reduces in size. (\cite{485752})

% from wiki:
%“All of the following numbers should be considered with caution: seemingly-minor changes to the rules of a game can change the numbers (which are often rough estimates anyway) by tremendous factors, which might easily be much greater than the numbers shown.”

The game-tree complexity (GTC) value itself relies on two approximate auxiliary values: the branching factor and the average game length. A possible formula to calculate GTC is as follows (\cite{enwiki:1129116028}):
% formula for GTC
\begin{align}
    GTC \ge log_{10} (b^d) 
    \label{eq:gtc}
\end{align}

where $b$ is the branching factor, and $d$ is the number of plies in an average game. It is important to note that game tree complexity is an approximate value. Small changes to either the calculation of the branching factor or average game length can affect the complexity value by a significant factor. Other possible formulae have been adapted to measure GTC based on the number of possible position options at a given ply, which may work better for games where the branching factor is continually changing (\cite{https://doi.org/10.48550/arxiv.1901.11161}). 

 The board size, state space complexity, and size of the feature space matrix are dimensionality estimates of intrinsic domain complexity. The GTC with the average game length and branching factor are measures of the extrinsic domain complexity and are OWL agent dependent.

In data science and perception domains, the environment space size (Figure \ref{fig:components}), or the upper bound is the number of pixels in imagery multiplied by the number of channels or the number of parameters, the number of classes in the dataset, and the number of pixel values. This is the size of the feature space matrix. Since these can be very big numbers, we use the ceiling of the logarithm to base 10 here as well. However, this formula does not take into account the task of the domain and the related constraints.
% formula for perception domain dimensionality
\begin{align}
    D_{upper bound} = log_{10} (\#pixels * \#channels * \#classes * \#pixelValues * \#images) 
    \label{eq:imageDim}
\end{align}
 
Dimensionality of a domain can be estimated using other methods, such as using covariance matrices of embedded time series signal and Gaussian random process (\citealt{Carroll2017}), or calculating fractal dimensionality when scale is an important factor (\citealt{flake2000computational}; \citealt{feldman2012chaos}).

%One of the ways that dimensionality can be calculated is described in the paper written by \cite{Carroll2017}. They describe it as a way to estimate embedding dimensions from a time series. Along with that, this estimation includes an estimate for the probability that the original estimate is correct. This estimation is done by comparing the "eigenvalues of covariance matrices created from an embedded signal to the eigenvalues for a covariance matrix of a Gaussian random process with the same dimension and number of points." Once calculated, a statistical test is used to give the probability that the eigenvalues for the embedded signal did not come from the Gaussian random process.

%having a large number of possible values across a small number of dimensions can be as difficult as having a small number of possible values across a larger number of dimensions, although depending on algorithm design they may have different impacts on performance. % this sound like density and dimensionality together
% Example of dimensionality estimation that could be used: Carroll, T. L., & Byers, J. M. (2017). Dimension from covariance matrices. Chaos: An Interdisciplinary Journal of Nonlinear Science, 27(2), 023101. https://doi.org/10.1063/1.4975063

% [How does State Transition Graph Size and dimensionality work over interactive AI domains? ]
% [How does Feature Space dimensionality and size work over non-interactive data tasks/domains?]
% [Dimensionality estimation with covariance matrix, and other methods to estimate dimensionality and size]
% Scale: ?

%-----------------------------------

\subsection{Sparsity} \label{5.2 sparsity}
 Sparsity relates to the rarity of successful strategies, or how thinly distributed is the information necessary to make successful decisions. This is related to how difficult it is to develop successful strategies and to perform well.  \cite{Hurley} describe sparsity of an information array as “small number of coefficients
contain a large proportion of the energy.”  Increased sparsity can increase bias. More sparsity implies less usable information which requires more complicated methods for solving a task optimally. This is why an understanding of sparsity and how it is measured is necessary for properly assessing the level of complexity of a domain. For intrinsic domain complexity (agent-independent) estimation we want to find the answer to the question ``How thinly spread out is the information?''.  For extrinsic (agent-dependent) the question we want to answer is ``What proportion of the possible attempts lead to success?''.

In classical AI action
%-and-planning 
domains, both the sparsity of paths that lead through the state space and the rarity of successful paths to the goal impact the complexity with regard to the task solution space and the performance space (see Figure \ref{fig:TTT_states}). When most states can be accessed through only a very small number of paths, the state space is more difficult to navigate and the domain is less forgiving of suboptimal decisions. It is easy to unintentionally enter a state from which the goal is no longer accessible. Significantly more memory or computation may be required to identify the optimal path correctly throughout the entirety of execution. 

In data science and perception domains, a sparsely distributed training dataset, especially when spread across a large feature space, significantly increases complexity with regard to the task solution space and performance. When there are many classes or prediction values that only have a few observations in the training data (i.e., many classes appear rarely in the training data) and when the training data in general is spread sparsely throughout a large dimensional feature space, there will be many modeling solutions with similar fit (given this minimal information), making it challenging to model the data distribution meaningfully in a way that supports accurate predictions. 

\cite{Hurley} describe six characteristics of a good sparsity measure. These are 
1) \emph{Robin Hood}: sparsity is decreased when data concentrated in smaller regions of the space are spread to larger regions of the space. 
2) \emph{Scaling}: scaling all of the data by a constant factor does not affect sparsity. 
3) \emph{Rising Tide}: adding a large value to all the data decreases sparsity. 
4) \emph{Cloning}: an exact copy of the data has the same sparsity as the original. 
5) \emph{Bill Gates}: as one piece of data becomes larger, sparsity is increased, and infinite value on one piece of data leads to maximum sparsity.
6) \emph{Babies}: works in the opposite way as the Bill Gates attribute. Introduction of new data with minimal value increases sparsity. The last two attributes (Bill Gates and Babies) work on the idea of adding maximum and minimum values to a set of data increasing the sparsity. 

Knowing these six conditions of a good sparsity measure, \cite{Hurley} studied different measures to see how well they satisfied each one of the six conditions of a good sparsity measure. They found that the \emph{pq-mean} $(p\leq1, q>1)$ and the \emph{Gini Index}  are the only measures to satisfy all six conditions. The \emph{Gini Index} is an indicator for when sources are separable, a property which itself relies on sparsity and it does not require the definition of a null value. Its equation is: 
% formula for Gini measeure of sparsity
\begin{align}
   S_{Gini} = 1-2\sum_{k=1}^{N} \frac{{{c}_{\left( k \right)}}}{{{\left\| {\vec{c}} \right\|}_{1}}}\left( \frac{N-k+\frac{1}{2}}{N} \right)  
     \text{for ordered data,  }{{c}_{(1)}}\le {{c}_{(2)}}\le ...\le {{c}_{(N)}} 
     \label{eq:Gini}
\end{align}
where $c$ is an array of information with $N$ elements. \cite{Hurley} noted that the \emph{Gini Index} was originally proposed in economics as a measure of the inequality of wealth and is still studied. In signal processing language this concept is referred to as “efficiency of representation” or “sparsity.” We considered and applied the \emph{Gini Index} as a measure of sparsity to the case study datasets described in section 6. 

%  H. Dalton, “The measurement of the inequity of incomes,” Econom. J., vol. 30, pp. 348–361, 1920.
% B. C. Arnold, Majorization and the Lorenz Order: A Brief Introduction. : Springer-Verlag, 1986.
% M. O. Lorenz, “Methods of measuring concentrations of wealth,” J. Amer. Stat. Assoc., 1905.
% C. Gini, “Measurement of inequality of incomes,” Econom. J., vol. 31, pp. 124–126, 1921.

%distribution of useful stuff across the problem, how many observations per concept 
% density estimation description from NLP and changing to fit our framework of components
% How dense/sparse are the observations over the problem space? 
% [How does State Transition Graph Sparsity work over interactive AI domains?]
% It can operate over all the components as input.
% [How does Feature Space Sparsity work over non-interactive data tasks/domains?]

%--------------------------------------

\subsection{Diversity} 
\label{sec:diversity}
We refer to diversity as the importance of information across the given domain components. Problems that include more diversity generally require more information and, therefore, more complicated solutions to address correctly. For intrinsic domain complexity (agent-independent) estimation we want to find the answer to the question ``How easy/difficult is it to summarize the information?''.  For extrinsic (agent-dependent) the question we want to answer is ``How high is the level of ambiguity in distinguishing good choices from bad ones?''.

Classical AI action problems have a very diverse state space (environment space) in which different states often have different possible action sets, where the same action may produce very different transitions, depending on the feature values of the current state. These action or planning problems have an increased complexity with regard to the task solution and the performance space as they require more information to navigate successfully. This often occurs in contexts in which a diverse set of skills is required to navigate the state space. Domains with high skill-set complexity will have high diversity.

In data science and perception domains, diversity in the feature space is when different features have very different properties with respect to the data distribution, for example, combining data from different sensors in cases with high skill-set complexity. Additionally, classes or prediction values, which reflect very diverse subgroups in the data, may be more difficult to model efficiently and accurately. 

Over the past half century statisticians, ecologists, economists and computer scientists proposed a variety of diversity measures. We distinguish between four types of diversity measures: variation, entropy, attribute, and distance measures. 

\textbf{Variation:} When the variation of the distribution of the task in a domain is higher, the diversity is higher, and so is the complexity level of the domain. We can capture variation with distribution. The most common measure of variation is the statistical variance, and it is the average square distance to the mean. 
The problem is, that the variance only works for numerical values.  

\textbf{Entropy:} The Shannon Entropy (referring to as entropy):
 $-\sum{{{p}_{i}}}\log_2 {{p}_{i}}$  
where $p_i$ is the probability of $i$ discrete random variables (events) of the task in the domain. Uniform probability yields maximum uncertainty and therefore maximum entropy of the task in the observed domain. With $i=N$ uniform probabilities $p$, the maximum Entropy will be $H_{max} = -log_2{{1/N}}$. In order to be able to compare entropies of different domains, we normalize the Shannon Entropy by dividing with the maximum entropy for that domain:
\begin{align}
     H_{normalized} = \frac{H}{H_{max}} = 
     \frac {-{\sum{{{p}_{i}}}} {log_2 {{p}_{i}}}} {\log_2{N}} % {-\log_2{\frac{1}{N}}} 
     \label{eq:EntropyNormalized}
\end{align}

The entropy measures the variability of the elements within a distribution. It is the only indicator that is continuous, monotonic, linear in the $p_i$ and supports properties of symmetry and additivity. It quantifies the number of questions one needs to ask to identify an element within a distribution: more questions, means more variability, which means higher complexity level of the domain. 
To take different types of entities into account we use entropy estimation. The entropy captures the evenness of a distribution across types. As the distribution is more even, there is higher entropy. The highest entropy in the domain is when the distribution is entirely even, and all the probabilities $p_i$ are equal. The highest diversity is however, when the entropy is between its domain maximum and zero value. When the distribution is maximally skewed the entropy will be close to zero. Therefore, lower entropy means a higher diversity level of the domain and the highest diversity level is when normalized entropy is close to 0.5. 

When calculating entropy, we need to pay attention to how we define probability $p$ and what bin size we choose. Different bin sizes of a distribution will result in different entropy. When comparing different domains, it is important to aware this fact and use the same principle of defining the probabilities and the same bin sizes for domains.  

Entropy is great for estimating diversity level, however, it fails to take into account the difference between the types. Evenly distributed similar types will have higher entropy than uneven distributed very different types.  For this reason, we want to use the statistical distance and attribute measures as well. 

\textbf{Statistical Distance Function:} The Distance function measures the assumed pre-existing statistical distance between pairs of types. The distance function estimates diversity by adding up all the distances between the members of the set and take an average. The more distance, the more diversity there is going to be. 

\textbf{Attribute:} Attribute measures are the second way to identify type level differences. This measure identifies the attributes of each type in the set, and then counts up the total number of unique attributes. More attributes implies a  higher diversity level, which in turns means a  higher complexity level.  
To take into account the different aspects of diversity we need to use the combination of these measures. 

%{\color{magenta} %The measure of diversity is variation, entropy, type, attributes, or the combination of these. 
%\begin{itemize}
   % \item Describe in general, in action domains and in data science and perception domains. 
    %\item Define the meaning of lower vs. higher entropy
   % \item from Cao, 2020: regarding diversity of source: entropy would reflect this information? Different sources => lower entropy?
%\end{itemize}}

%-----------------------------------------
%\subsection{Ambiguity} Maybe for next time? 
%Task-dependent difficulty metric, determining the difficulty of distinguishing goal/non-goal (ie, success vs failure) paths in a state transition graph, and the difficulty of distinguishing between classes in a data science problem. 

\subsection{Agent-Based Extrinsic Measures}

\begin{figure}[htb]
  \centering
  \begin{minipage}[b]{0.45\textwidth}
    \includegraphics[width=\textwidth]{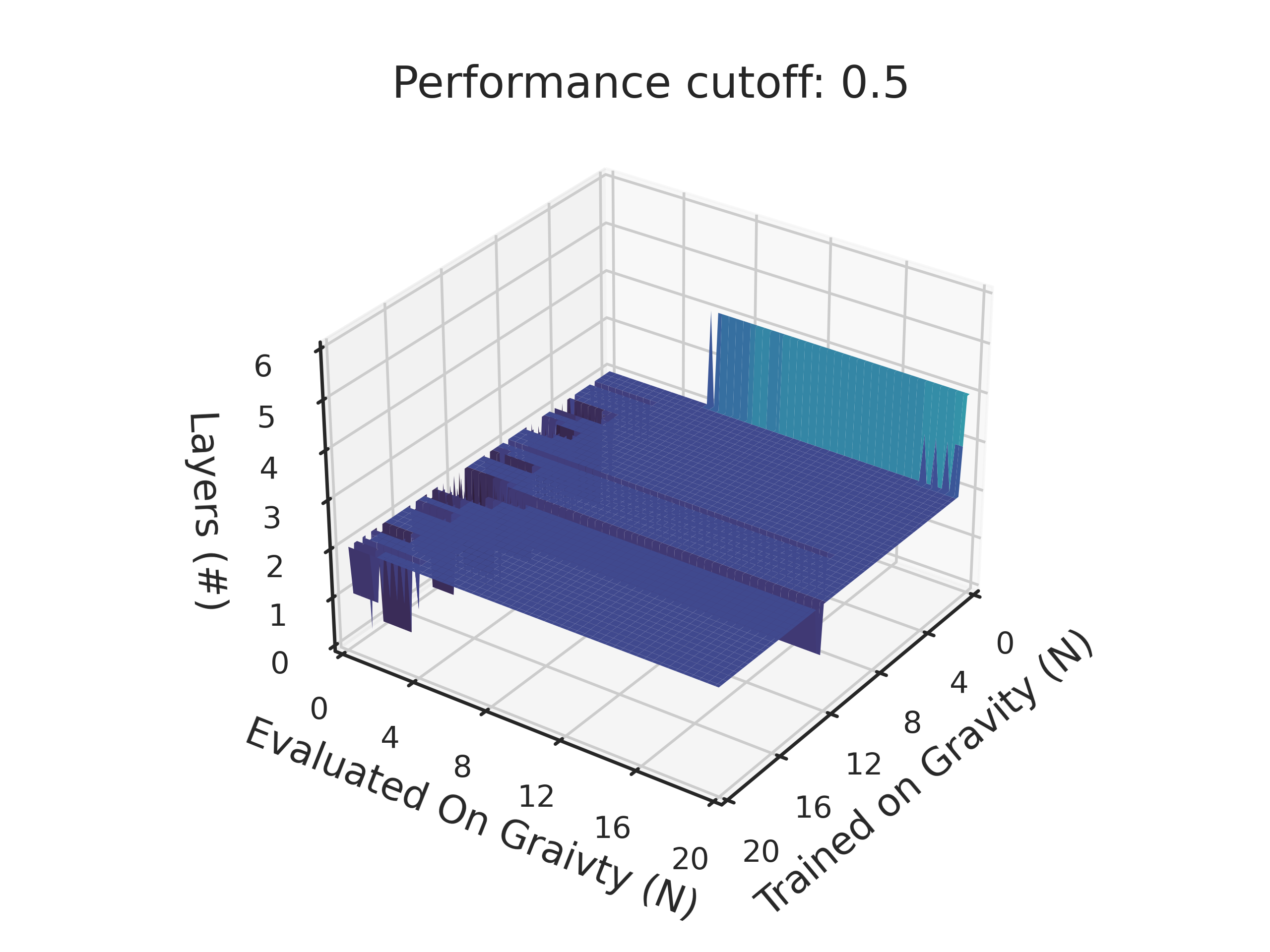}
  \end{minipage}
  \begin{minipage}[b]{0.45\textwidth}
    \includegraphics[width=\textwidth]{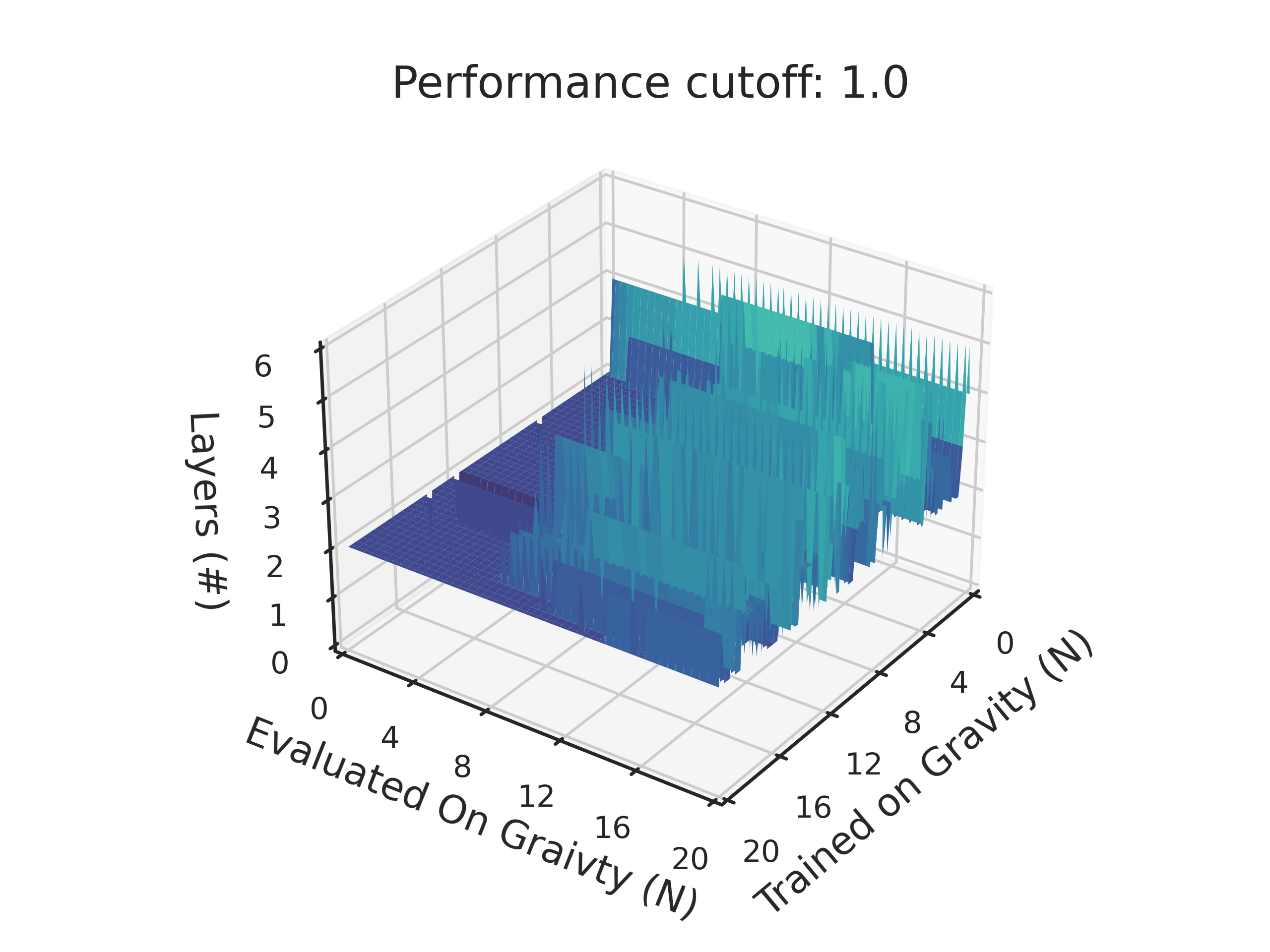}
  \end{minipage}
  \caption{ Minimum number of layers needed by DQN agent to achieve (left) 50\% performance (pole balanced for 100 seconds) and (right) 100\% performance (pole balanced for 200 seconds) on tasks in which the agent is trained on one value of gravity but is tested on another value. } \label{fig:cartpole_min_layers}
  \end{figure}

The above dimensions define the spaces in which an agent resides in order to solve tasks in the domain. One way to identify relationships across these spaces is to define a class of agents from the skills space to solve a task and to compare the minimal complexity of an agent from the class that is capable of achieving different points in the performance space. \citeauthor{Pereyda2020} (\citeyear{Pereyda2020}) proposed a measure of task complexity that is defined as the sum of complexities of minimal-complexity agents capable of achieving the possible range of performance scores for the task. For example, Figure~\ref{fig:cartpole_min_layers} depicts the complexity of a deep Q-learning agent in terms of the number of layers necessary to achieve half (0.5) and full (1.0) performance on the Cartpole task (balancing a pole on a cart by pushing the cart left and right), in which the agent is trained on one setting of gravity and is tested on another (to mimic the open-world novelty of a change in gravity).  As the figure shows, more complex agents are needed to achieve higher performance, especially on tasks for which gravity increases, but less so when gravity decreases. These agent-based extrinsic results allow us to evaluate the complexity of tasks and provide insights into the intrinsic domain complexity.

%\begin{figure}[htb]
%  \centering
%  \includegraphics[width=10cm]{figures/cartpole.jpg}
%  \caption{ Minimum number of layers needed by DQN agent to achieve %(left) 50\% performance (pole balanced for 100 seconds) and (right) 100\% performance (pole balanced for 200 seconds) on tasks in which the agent is trained on one value of gravity but is tested on another value. } \label{fig:cartpole}
%\end{figure}

%------------------------------------------
%\paragraph{} {\color{magenta} This paragraph might need to go somewhere else? or be rewritten? } Domain complexity level in a interdisciplinary light differs in different ways. Domain complexity level can indicate that the domain is simple, complicated, or complex in different ways. For this reason, we propose using a final, single value of a complexity level rather than complexity level values for the environment, policy, solution, and planning spaces described in Section 3. These four spaces' complexity level can be expressed as a bar chart or a spider chart (with its caveats). The skills space expressed in these four spaces would show what skills are needed for an agent to perform a task successfully in that domain. 

%High level of complexity does not necessary mean that the domain is a complex system. 

%== 8 ===================================================
\section{Case Study Results}
\label{Case studies}
Here we demonstrate the domain complexity level estimates on several cases from action, perception, and data science domains.

%----- Tic-Tac-Toe and Qubic --------
\subsection{Tic-Tac-Toe and Qubic (Action Domains)}
Tic-tac-toe is a two-player game on a 3x3 grid where players take turns marking the spaces with an $X$ or $O$. A player wins by filling in a complete row, or line. There are eight possible winning lines. Winning lines are three horizontal lines, three vertical lines, and two diagonal lines. The game can also result in a draw, with all nine spaces being filled with no winning row. While traditional tic-tac-toe is in two dimensions, Qubic is a three-dimensional version of tic-tac-toe with a 4x4x4 grid. Qubic can be played by three players, but for the purpose of this study, the focus was on two-player games. Qubic is a more complex game with seventy-six possible winning combinations. On each of the four planes, there are 4 horizontal, 4 vertical, and 2 diagonal winning lines. Spanning all four planes, there are 8 diagonal-vertical, 8 diagonal-horizontal, 16 straight-line, and 4 diagonal winning lines. Qubic can also end in a draw.

\paragraph{\textbf{Dimensionality}}
Both games are classic examples of combinatorial games, where players alternate turns until a player wins or the game ends in a draw. For tic-tac-toe, an upper bound of the state-space complexity is given by $3^9 = 19,683$ transitions, which represents the three possible options for each cell of the 3x3 board during each play - $X$, $O$, or empty. Similarly with Qubic, an upper bound is given by ${{3}^{64}}=3.43\,x\,{{10}^{30}}$
%$3^64 = 3.4336838e+30$ 
transitions. However, these upper bounds include illegal plays (i.e., games where the board is filled in only with X's, or games with uneven X's and O's given the amount of turns played). Since these games are combinatorial, the formula below utilizes combinatorics to find the approximate state-space complexity (SSC) of both games (\cite{613505}):
% formula for state space
\begin{align}
    SSC = \sum_{i=1}^{P} {N^D \choose \lfloor \frac{i+1}{2} \rfloor} \times {N^D - \lfloor \frac{i+1}{2} \rfloor \choose \lfloor \frac{i}{2} \rfloor}
    \label{eq:TTT_state_space}
\end{align}
where $N$ is the dimensions of a single board (3 x 3, for example), $D$ is the dimensions of the entire board (2-D, for example), $i$ is the ply (or turn) of the game, and $P$ is the total number of plies (states). $\lfloor \frac{i+1}{2} \rfloor$ represents Player 1 (typically X), and $\lfloor \frac{i}{2} \rfloor$ represents Player 2 (typically O). For each turn or ply, the formula calculates the number of possible position combinations between the players. For example, if looking at the second ply, then X has one position on the board, and there are 8 remaining spots for O to possibly fill.

This formula only provides an approximation of the number of state transitions, as there are still a few instances of illegal plays included in the combinations. Some of the combinations include games were an extra position is filled, even after the game has ended. However, the approximation is close enough to provide a state-space complexity value. Taking the base-10 log of the total number of state transitions provides complexity values of 3 and 30 for tic-tac-toe and Qubic, respectively.

The game tree complexity of tic-tac-toe can be computed by assuming that the average game length is 9 plays. The majority of tic-tac-toe games end in a tie, thus 9 total moves. The branching factor changes with each turn of a player, starting from 9 to 8 to 7, and so on. With a branching factor of $9 - i$ at level $i$, there are $9! = 362880$ terminal nodes. Taking the base-10 log of this value gives the game tree complexity value of 5.
Similarly with Qubic, we assume an average game length of 20 moves, and the branching factor is $64 - i$ at level $i$. Since the average length of a game of Qubic is not the same as the total number of possible plays allowed (64), the game tree complexity is calculated as $\frac{64!}{44!}$. We divide by $44!$ to eliminate the plays longer than 20 moves. Since $\frac{64!}{44!} \approx 10^{34}$, the game tree complexity value for Qubic is 34.

\paragraph{\textbf{Sparsity}}
Board games, such as tic-tac-toe and Qubic, are played on a matrix-like board where players fill in a square or point. For an action space such as a board game, we can consider the number of pieces on the board as well as their proximity to each other. (\cite{inproceedings}) As shown with Equation \ref{eq:Gini}, it is possible to evaluate an array of information with N elements. To utilize this formula, a matrix would be created for each possible position on the board (for either tic-tac-toe or Qubic) to calculate the sparsity. (\cite{Hurley}) However, in practice, this does require a large number matrices to be created. For tic-tac-toe, nearly 6,000 possible positions would have to be accounted for. For Qubic, the number is even higher.

\paragraph{\textbf{Diversity}}
We calculated the entropy for the tic-tac-toe from the number of unique states in each ply of the state transition space using equation \ref{eq:EntropyNormalized}. The normalized entropy for tic-tac-toe is 0.3959. It shows the diversity level of unique moves that can be taken in the game.

%----- CartPole --------
\subsection{CartPole 2D and CartPole 3D (Action Domains)}

\begin{figure}[htb]
  \centering
  \includegraphics[width=6cm]{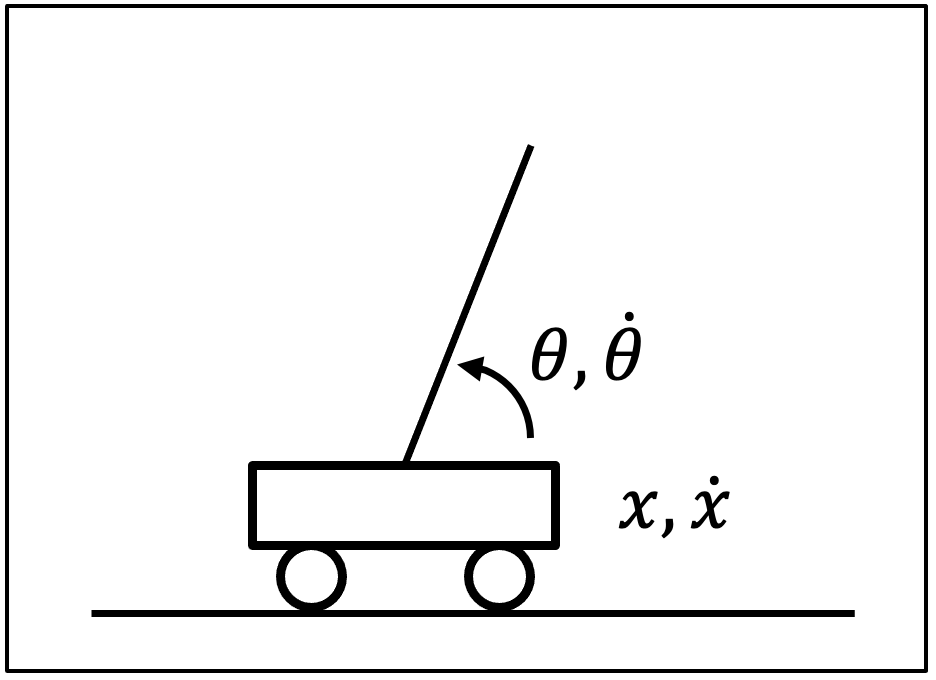}\hspace{0.25in}
  \includegraphics[width=6cm]{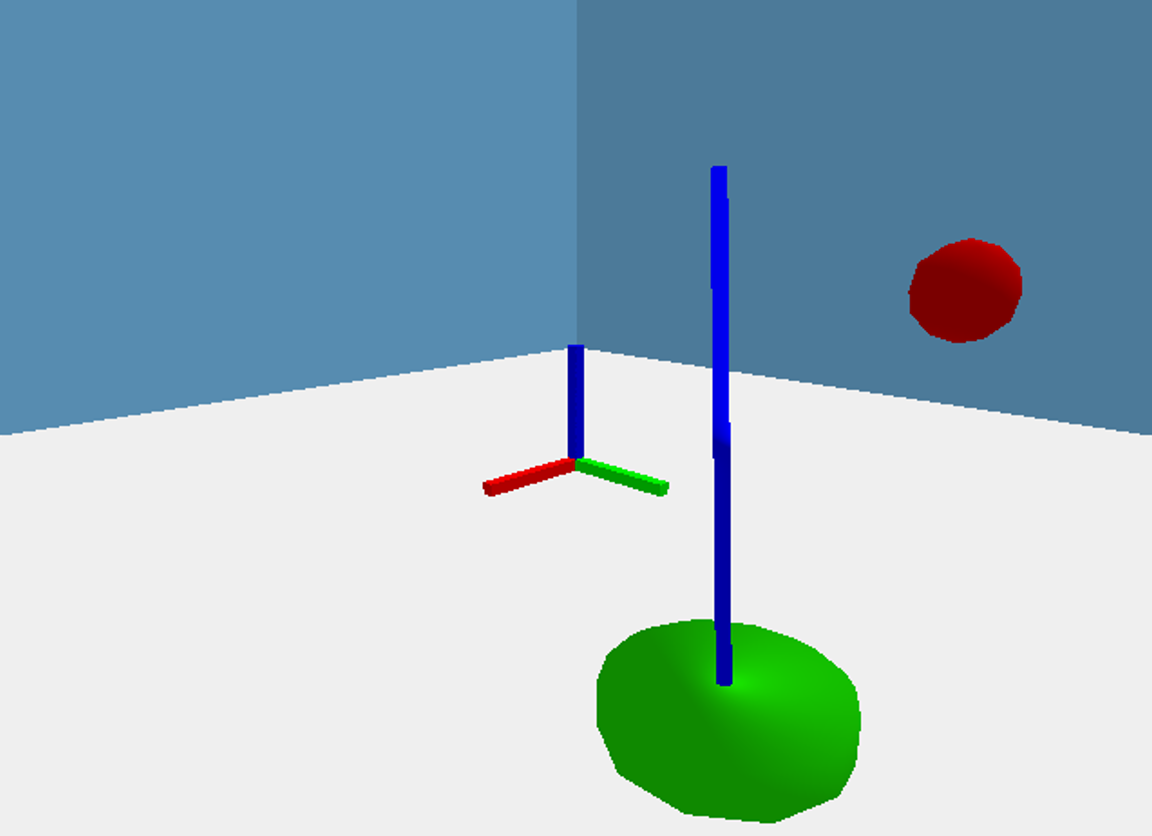}
  \caption{The classic two-dimensional CartPole domain (left) and the three-dimensional version (right).}
  \label{fig:cartpole}
\end{figure}

The classic two-dimensional CartPole domain \cite{openaigym} (see Figure~\ref{fig:cartpole}) consists of a cart that moves left and right along a track with a pole attached to the cart that can swing left and right. The agent can push the cart left or right with the goal of keeping the pole balanced. The CartPole2D domain typically provides the cart’s position $x$ and velocity $\dot{x}$ along with the pole’s angle $\theta$ and angular velocity $\dot{\theta}$. The CartPole2D world model further defines a set of parameters for each instantiation of the world. The cart can vary in size $S$ and mass $M_c$, the track can vary in length $L_t$, the pole can vary in length $L_p$ and mass $M_p$, gravity $G$ and track friction $F_t$ can vary, and the push force $F_p$ can vary.

The three-dimensional CartPole domain \cite{boult-algs-2022} (see Figure~\ref{fig:cartpole}) increases the complexity over the CartPole2D domain by allowing the cart to move in two dimensions, the pole to move in three dimensions, and the introduction of balls that move around in the three-dimensional environment. The agent can push the cart in the $\pm x$ and $\pm y$ directions. The CartPole3D domain provides the cart’s position $(x, y)$ and velocity $(\dot{x}, \dot{y})$, the pole’s Euler angles (yaw $\theta_y$, pitch $\theta_p$, roll $\theta_r$) and angular velocity $(\dot{\theta}_y, \dot{\theta}_p, \dot{\theta}_r)$, and each ball’s position $(x_i, y_i, z_i)$ and velocity $(\dot{x}_i, \dot{y}_i, \dot{z}_i)$. The CartPole3D world model parameters include the cart’s size and mass, the pole’s length and mass, ball size and mass, world dimensions, gravity, floor friction, and push force.

\paragraph{\textbf{Dimensionality}}

The game complexity measures (\cite{enwiki:1129116028}) are designed for the analysis of static, discrete games (e.g., board games), so their application to continuous, real-time domains requires some modification. First, there is the obvious need to discretize the continuous attributes of the domain. For example, the CartPole2D cart position $x$ over a track length $L_t=10$ can be defined in terms of discrete positions \{-5, -4, -3, -2, -1, 0, 1, 2, 3, 4, 5\}. The choice of precision is somewhat arbitrary, but is guided by the need to distinguish world states relevant to the domain’s main task goal of balancing the cart. Lower precision (e.g., \{left, middle, right\}) may render the control problem impossible, while higher precision (e.g., \{-5.000, -4.999, -4.998, $\ldots$\}) may result in redundant subsets of states that are indistinguishable from each other. The second modification involves the dynamic nature of real-time worlds, where the objects in the world are moving continuously and the agent is interacting with the world continuously. So, there is no notion of a ``ply'' in which the world remains static while the agent deliberates. In simulated real-time domains, this issue is typically addressed by advancing the world for a fixed amount of time (i.e., ``game tick'') and then pausing the world while the agent deliberates and returns a discrete action. This discretization of the world-agent interaction impacts the definitions of game-tree complexity and average game length, since these measures require a discrete transition between states.

The third modification to the game complexity scenario when applied to real-time domains is the definition of terminal states. In a board game, the terminal states are typically those that result in a win or loss for the agent. While a real-time game can have well-defined terminal states (e.g., the pole angle has exceeded some threshold), the CartPole domains typically have a fixed time limit, such that if the time limit is exceeded with the pole still balanced, the agent ``wins'' and the game terminates. But the goal could also be defined as keeping the pole balanced indefinitely. This terminal state issue further complicates the definitions of game-tree complexity and average game length, since these measures require a well-defined set of terminal states.

Given the above qualifications for real-time domains, we can apply the game complexity measures to the CartPole domains. For CartPole2D, we make the following assumptions about the domain’s definitions. Items 1-7 pertain to a particular instance of the CartPole2D domain, and items 8-11 pertain to the definition of a world state. 

\begin{enumerate}
\item Cart size and mass each range over 10 possible values.
\item Pole length and mass each range over 10 possible values.
\item The push force ranges over 10 possible values.
\item Gravity and friction each range over 10 possible values.
\item The length of the track ranges over 10 possible values.
\item There are 2 actions that are always applicable.
\item The maximum game time is 100 game ticks.
\item The cart’s position $x$ ranges over 100 possible values.
\item The cart’s velocity $\dot{x}$ ranges over 10 possible values.
\item The pole’s angle $\theta$ ranges over 100 possible values.
\item The pole’s angular velocity $\dot{\theta}$ ranges over 10 possible values.
\end{enumerate}

Based on these assumptions, we can compute the game complexity measures for CartPole2D, as shown in Table~\ref{tab:cartpolestats}. Board size has no real analogue in real-time games, but we can estimate the board size as the number of different positions of the cart ($10^2$), assuming the pole angle and both velocities are initially zero. A state is defined by the position and velocity of the cart and pole $(x, \dot{x}, \theta, \dot{\theta})$. Thus, the number of states is $100 x 10 x 100 x 10 = 10^6$, and the state-space complexity is $\log_{10} 10^6 = 6$. The space of game trees is bounded by the number of actions (2) at each game tick, the minimum number of game ticks leading to a loss (pole angle out of range), and the maximum number of game ticks leading to a win (100). The minimum number of game ticks depends on the values of the initial state and the capabilities of the agent, but an immediate loss is possible, which results in a minimum game ticks of 2. Thus, the space of game trees can be defined as $\sum_{i=1}^{100} 2^i = 2^{101} - 2$. Thus, the game-tree complexity is $\log_{10} (2^{101} - 2) = 30.4$. The average game length also depends on the initial state and the capabilities of the agent, but we can estimate the average game length as half of the maximum game ticks, or 50. Finally, the branching factor is fixed at 2.

The above measures apply to a particular instance of the game world, but we are also interested in the space of games, since an agent that adapts to novelty can encounter any of these possible games. Assuming the number of actions and game time is fixed, we can define the game-space complexity as the product of items 1-5 above, which is $10^8$. If we further consider the possible initial states, then there is an additional factor of $10^6$, for a game space size of $10^{14}$. Using the log base 10 for this measure as well, the game space complexity = 14.

\begin{center}
\begin{table}
\caption{Game complexity measures for CartPole2D and CartPole3D domains.}
\label{tab:cartpolestats}
\begin{tabular}{|l|r|r|} \hline
{\bf Measure} & {\bf CartPole2D} & {\bf CartPole3D} \\ \hline
Board size & $10^2$ & $10^10$ \\ \hline
State-space complexity & 6 & 24 \\ \hline
Game-tree complexity & 30 & 60 \\ \hline
Average game length & 50 & 50 \\ \hline
Branching factor & 2 & 4  \\ \hline
Game-space complexity & 14 & 27 \\ \hline
\end{tabular}
\end{table}
\end{center}

For CartPole3D, we make the following assumptions about the domain’s definitions. Items 1-9 pertain to a particular instance of the CartPole3D domain, and items 10-15 pertain to the definition of a world state. 

\begin{enumerate}
\item Cart size and mass each range over 10 possible values.
\item Pole length and mass each range over 10 possible values.
\item Ball size and mass each range over 10 possible values.
\item The number of balls has 5 possible values (0-4).
\item The push force ranges over 10 possible values.
\item Gravity and friction each range over 10 possible values.
\item The size of the world ranges over 10 possible values.
\item There are 4 actions that are always applicable.
\item The maximum game time is 100 game ticks.
\item The cart’s position $(x, y)$ each range over 100 possible values.
\item The cart’s velocity $(\dot{x}, \dot{y})$ each range over $10 x 10$ possible values.
\item The pole’s angles $(\theta_y, \theta_p, \theta_r)$ each range over 100 possible values.
\item The pole’s angular velocities $(\dot{\theta}_y, \dot{\theta}_p, \dot{\theta}_r)$ each range over 10 possible values.
\item Each ball’s position $(x_i, y_i, z_i)$ each range over 100 possible values.
\item Each ball’s velocity $(\dot{x}_i, \dot{y}_i, \dot{z}_i)$ each range over 10 possible values
\end{enumerate}

Based on these assumptions, we can compute the game complexity measures for CartPole3D, as shown in Table~\ref{tab:cartpolestats}. We can estimate the board size as the number of different positions of the cart ($100^2$) and a single ball ($100^3$), assuming the pole angles and all velocities are initially zero. A state is defined by the position and velocity of the cart, pole and balls $(x, y, \dot{x}, \dot{y}, \theta_y, \theta_p, \theta_r, \dot{\theta}_y, \dot{\theta}_p, \dot{\theta}_r, x_i, y_i, z_i, \dot{x}_i, \dot{y}_i, \dot{z}_i)$. 

Assuming one ball, the number of states is $100^2 x 10^2 x 100^3 x 10^3 x 100^3 x 10^3 = 10^{24}$, and the state-space complexity is $\log_{10} 10^{24} = 24$. The space of game trees is bounded by the number of actions (4) at each game tick, the minimum number of game ticks leading to a loss (pole angle out of range), and the maximum number of game ticks leading to a win (100). The minimum number of game ticks depends on the values of the initial state and the capabilities of the agent, but an immediate loss is possible, which results in a minimum game ticks of 2. Thus, the space of game trees can be defined as $\sum_{i=1}^{100} 4^i = (4^{101} - 4) / 3$, and the game-tree complexity is $\log_{10} (4^{101} - 4) / 3) = 60.3$. The average game length also depends on the initial state and the capabilities of the agent, but we can estimate the average game length as half of the maximum game ticks, or 50. Finally, the branching factor is fixed at 4.

For game space complexity, assuming the number of actions and game time is fixed, we can define the game-space complexity as the product of items 1-7 above, which is $4 x 10^{10}$. If we further consider the possible initial states, then there is an additional factor of $4 x 10^{16}$, for a game space size of $1.6 x 10^{27}$ and game space complexity $\log_{10} 1.6 x 10^{27} = 27.2$.

For our comparison of sparsity and diversity we are introducing a variation on the Cartpole2D domain which we are labeling Cartpole2D-G. Cartpole2D-G is the exact same as Cartpole2D, except for a significant increase in the gravity of the domain. We are increasing gravity from the standard $9.8 \frac{m}{s^2}$ to $250.0 \frac{m}{s^2}$. Previous experimentation has shown an approximate $50\%$ decrease in average performance of a DQN based solution between Cartpole2D and Cartpole2D-G.

\paragraph{\textbf{Sparsity}}

\paragraph{}

Analytical : We can calculate the sparsity of a domain by calculating the space of available action sequences that result in a success, as defined by the domain. For most domains this is largely incalculable, but for Cartpole this is straightforward. We can determine the maximum number of consistent constant actions (e.g., how many push left actions in a row) before the pole falls over for each variation. This results in the maximum number of constant actions that can be applied while remaining in the solution path. For the three domains Cartpole2D, Cartpole2D-G, and Cartpole3D we found the constant action limit to be 9.37, 9.22, and 10.6 respectively. We can calculate the solution space for each of these domains by considering the action limit and actions that would in a successful game (e.g., pole remains balanced for 200 time steps). For Cartpole2D there are only two opposite actions, so an action either moves the path one step closer or one step further from the solution space. For Cartpole3D this same idea is applied, however in two directions. While the orthogonal actions are not independent of each other, the approximation is still useful for determining sparsity. The size of the solution space is then divided by the total number of possible paths to determine the sparsity of the domain. The sizes of the solution space and games space was determined through a Monte Carlo approximation. This approximation along with our constant action limits were combined to generate the final sparsity values in Table \ref{table:cartpole_sparsity}. Here we see that the sparsity of Cartpole2D-G is slightly lower than CartPole2D-G, while both are significantly higher than CartPole3D.

\begin{table}[]
\centering
\caption{Sparsity table using a Monte Carlo approximation to estimate the set of solutions divided by the set of possible spaces, with our experimentally determined action limit.}
\begin{tabular}{|c|c|c|c|}
\hline
         & Cartpole2D & Cartpole2D-G & Cartpole3D \\ \hline
Sparsity & 0.1171    & 0.1118      & 0.054    \\ \hline
\end{tabular}
\label{table:cartpole_sparsity}
\end{table}

\paragraph{}

\begin{figure}[h]
\centering
\includegraphics[width=\textwidth]{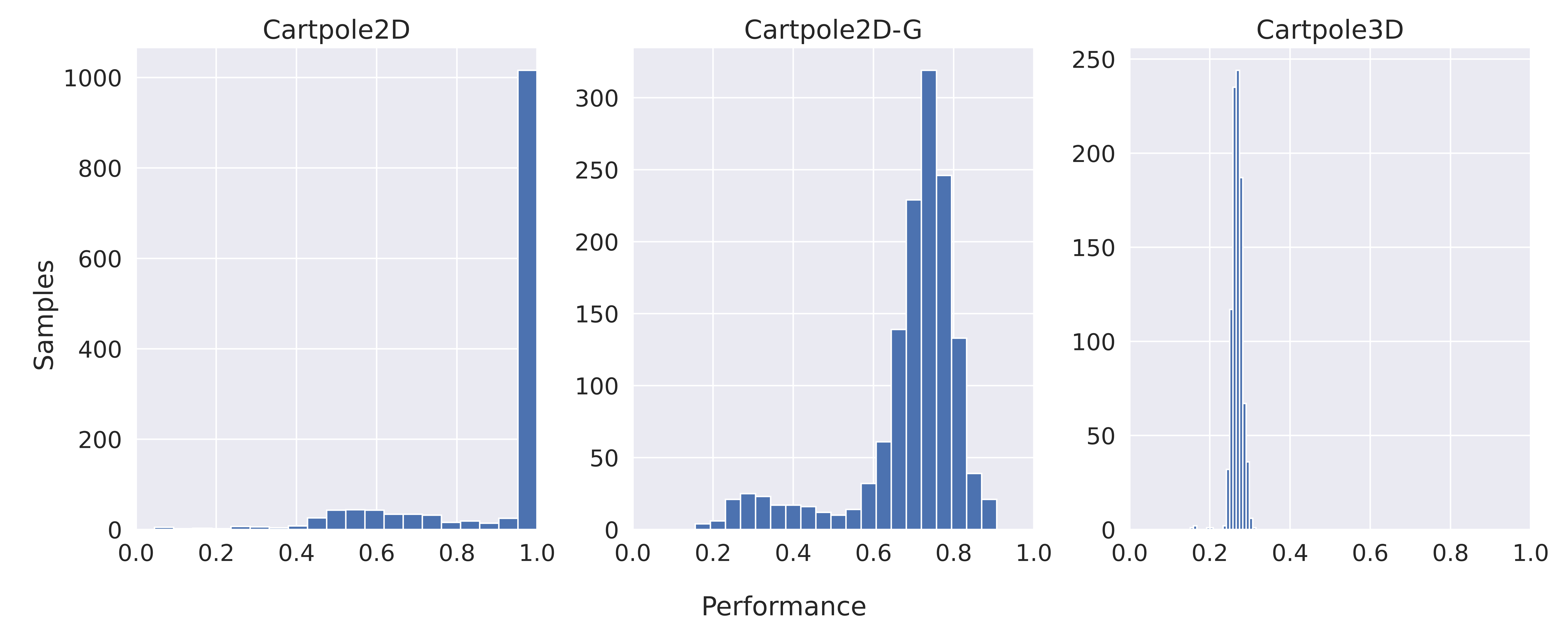}
\caption{Distribution of trained DQN agents for fixed training time. We sampled approximately 1400 agents for each domain. The Gini Coefficent for Cartpole2D, Cartpole2D-G, and Cartpole3D are 0.092, 0.097 and 0.026 respectively.}
\label{fig:cartpole_histogram}
\end{figure}

Experimental: For this method we train a large population of DQN based AI systems. The shapes of these agents are thirty-two nodes across six layers, and the corresponding input and output sizes based on the domain used. Each agent was trained for a fixed number of domain interactions (100,000) instead of waiting until convergence. The goal of this method is to demonstrate the probability of finding a certain performance achieved given a fixed time. With this fixed length searching, we can generate a population of performances and apply the Gini Equation \ref{eq:Gini}. We believe this to be analogous to the definition of sparsity previously discussed. The results of this experimentation can be observed in Figure \ref{fig:cartpole_histogram}. We observe that the Gini-based sparsities for CartPole2D and CartPole2D-G are similarly close, while the sparsity for CartPole3D is significantly lower.

\paragraph{\textbf{Diversity}}
We measure the diversity of the CartPole domains using entropy, which is calculated as follows. We gathered 20,000 samples of every feature-action pairing for all three domains. This was done using random action sampling within the selected domain. We then normalized and discretized the feature data into 256 bins for each feature in the domain. This creates a probability distribution over each individual feature and over the actions. We then summed the resulting Shannon entropy over all features and separately for the actions. The resulting (feature, action) The entropies for the domains are (20.556, 0.999), (17.626, 1.0) (99.999, 2.322), for Cartpole2D, Cartpole2D-G, and Cartpole3D respectively. Again, we see similar values for CartPole2D and CartPole2D-G, but much higher diversity for CartPole3D.

%----- Monopoly --------
\subsection{Monopoly (Action Domain)}
Monopoly is a zero-sum game of four players, where the goal is to acquire ‘properties’ on the board in a strategic manner. The game was first patented in 1935, and has been published in many versions around the world. The rules are similar across all of these versions. While there is considerable stochasticity in the game (players move around by rolling dice in turn), it is also a test of decision making under conditions of uncertainty. Because it has many small details, a realistic simulator for Monopoly was lacking until recently.

Similar to Poker and other games that are played under uncertainty, in Monopoly (i) players making sub-optimal decisions can win the game due to luck, (ii) optimal decision-making is no guarantee of a win. Indeed, even in tournament versions of Monopoly, well-meaning strategies yield 50-80\% probability of winning, compared to an average of 25\% (which is necessarily the case in a 4-person zero-sum game with one winner). In terms of the six space components:

Environment space: Monopoly has a non-trivial number of entities and complex phenomena. In addition to the four players, there is also a Bank in the background that originally owns all the properties and gives players a constant amount of money each time they pass Go. There are also properties, houses, hotels, chance and community cards, a jail, and dice. Some of these entities are governed by stochasticity. The rules and relations governing how these entities interact are complex, and in the original game, hard to always remember. Events and goals can play out over long time horizons. For example, a player may make certain moves with a goal in mind that only materializes (with some probability) after a certain number of moves. 

Task solution space: Monopoly has a large task solution space even at an intuitive level. There are multiple routes to winning a game, although some are more probable than others. What makes the task solution space particularly complex is its stochastic nature: there is no guarantee of a win, even when a hypothetical agent is playing well against other ‘worse’ players. 
Because there are three ‘other’ players involved, the extrinsic domain complexity of Monopoly is arguably higher than its intrinsic complexity, when the three other players can be treated as environmental entities, similar to the more static properties and cards. The skills needed for Monopoly include decision making, cooperation, understanding of tradeoffs, temporality (e.g., should I trade a property now with another player or wait?) and high adaptiveness (depending on the behavior of the other players). However, Monopoly does not require physical skills in a computational environment, or even a knowledge of physics. The other spaces (goal, performance and planning) are complex because of their interconnectedness to the other three players. Against relatively simple and non-adaptive opponents, an agent may not have to plan as carefully, and manage to win even within a limited performance space. Against highly adaptive agents, the spaces expand significantly. 

In evaluating these measures for Monopoly, we adopt commonsense definitions and assumptions, where applicable. Because there are many small details in the game, we focus on the main elements of the game for our calculations.   

\paragraph{\textbf{Dimensionality}}
Although the board size for Monopoly is only 40 (as there are 40 unique positions on a Monopoly gameboard), the environment space complexity is very high and can only be approximately calculated as follows. First, ignoring all other parameters, we know that each of the four players can occupy any one of the 40 board positions at a given point of time. Hence, there are $40^4$ or 2.56 $\times 10^6$ environmental states even if all rules and events are ignored. Second, we consider the `ownership' of properties on the Monopoly board. There are 28 properties (including railroads and utilities) that can be owned. Note, however, that properties can be owned by the Bank (which is another way of saying they are owned by none of the players). Since any one of five entities (the four players and the Bank) must own a given property at a given point of time, there are $ 3.73 \times 10^{19} \approx 5^{28} $ ownership-based environmental states. Ownership of properties is independent of where the players are located; hence, using only these two aspects of the game, we obtain $9.5 \times 10^{25}$ environmental states. Third, we take into account the cash owned by the players. In most modern editions of the game, the total amount of money that is available to be distributed (or in the bank) is $\backsim 20,000\$$. To calculate an upper bound, we assume that a player could own between 0 and 20,000. We also have to take the Bank into account as a fifth money-owning entity. This yields $\backsim 20,000$ states for each of the five entities, yielding an upper bound of $20,000^5=32 \times 10^{20}$ states. Multiplying by $9.5 X 10^{25}$ we obtain $3.04 X 10^{47}$ states. This is still not an upper bound since we have not taken into account `minor' aspects of the game such as ownership of the `get out of jail free' cards, improvements of properties with houses and hotels, mortgage status of properties, and bankruptcy status of players. These considerations are not independent of the three aspects above. However, the most important of these affecting the environment space are the improvement status of properties and the mortgage status of properties. To calculate the latter, consider that each of the 28 properties can be mortgaged or unmortgaged. Hence, there are $2^{28} = 2.7 X 10^8$ mortgage-related states. Considering improvement, an obvious upper-bound calculation is that each one of the 22 real-estate properties could be (i) unimproved, (ii) improved with 1,2,3 or 4 houses, (iii) improved with a hotel. Hence, there are six states possible for each of these properties, yielding $6^{22} = 1.32 X 10^{17}$ states. Multiplying all of this, we find that the number of possible environment states well exceeds $10^{72}$. Taking minor aspects of the game into account, a reasonable upper bound of the environment space for the full game is $\backsim 10^{80}$. 

The task solution space for Monopoly is similar to the environment space: a good upper bound assumption is that each of the environment space positions is, in fact, reachable from the initial position when all players are at Go with 1500 in starting cash, and with all ownable properties owned by the Bank. 

Unlike the environment space complexity, it is more feasible to calculate a prototypical (most likely to occur during much of the game) for the branching factor in Monopoly. To do so, we rely on the dice. Suppose we are in an environment space E. A single player will roll the dice on the next move. There are 12 possibilities (11 dice sums, but the player could be in jail, and unable to move). Hence, the branching factor is simply 12. 
The reason that this is somewhat of a `prototypical' bound is that players could trade with one another, buy and improve properties, engage in auctioneering, and so on. These would also be represented as branches because they will potentially change the game state (although not always e.g., a state will not change if a player offers a trade to another player, but the second player refuses; hence, nothing would change as a consequence). In principle, a player could offer many trades, improve properties etc. in a single move. In practice, there is usually a limit, and this limit is far below 12. Hence, the best branching factor to assume for Monopoly is 12.  

Determining the average game length is an empirical question for Monopoly as the theoretical upper bound on game length is infinite. According to \cite{AGLMonopoly1}, an average game lasts around 30 turns per player, which would yield a total of 120 turns. Each `turn' has the potential to change the current environment state. Therefore, from a game-tree standpoint, the average game length is 120 `edges' in the tree. An alternative way is to calculate the average number of dice rolls per game (which is around 200 per \cite{AGLMonopoly2}) but we believe that the former is a better estimate of average game length.

\paragraph{\textbf{Sparsity}}
There are two viable ways to determine the level of `sparsity' in Monopoly. If we represent Monopoly as a matrix, where environment spaces are on the two dimensions of the matrix, and we query the probability of going from one (arbitrarily selected) environment space to another arbitrarily selected environment space, the answer would generally be near-zero if we assume that all players are playing with the goal to win. 

A second, less restrictive, method is to assume that the `agent' can change its goal: instead of getting to a winning state, the agent may be trying to get to an arbitrarily selected environment state as its newly selected goal. If we assume that the game has been played for a few turns and that the agent is in the lead (which would have around 25\% probability if the agents are evenly matched), there is some probability that this `winning' agent has more control over the board and is in a better negotiating position than the other agents. An upper bound, therefore (with evenly matched agents), is 25\%. The sparsity of Monopoly would then range from 0 to 25\%, but in a realistic game would be much closer to 0. Intuitively, this is because a non-trivial part of the game is determined by elements that are outside the agent’s control, including the other agents' strategies and the stochastic elements in the environment (the roll of the dice). Once a few die rolls have occurred, and all the players have taken at least one full journey around the board, the set of paths through the game-tree to winning (for any given player) is significantly constrained, leading to naturally sparse solutions. 

\paragraph{\textbf{Diversity}}
Entropy is the best measure of computing complexity of Monopoly along the dimension of diversity. To calculate the entropy of Monopoly from an agent-dependent perspective, we assume a `fixed' agent that represents a bundle of strategies, playing against other agents with bundles of strategies. If we assume that all agents are heterogeneous and competitive, and that our `fixed' agent does not have any competitive a priori advantage, then the zero-sum nature of Monopoly ensures an entropy of 1.0. Technically, the game is not purely zero-sum as the `order' of the players (e.g., the first player to roll the dice has an advantage) can make a small difference. But since the order is also determined randomly, the game still becomes zero-sum overall.

However, suppose that we do have a competitive advantage. For instance, we pursue established winning strategies like owning all railroads (possibly through trade) or owning both Boardwalk and Park Place. The former strategy boosts our chance of winning to 80\% while the latter boosts our chance to 50\%. If we assume that the other three players divide up the `rest' of the winning probabilities equally among themselves, the former strategy yields an entropy 0.52, while the latter strategy yields an entropy of $\backsim 0.9$. In either case, the entropy tends to be high, even with a strong competitive advantage. This illustrates the high complexity of Monopoly even when some agents perform better than others. A win is not guaranteed. 

Furthermore, if we calculate the entropy of Monopoly from an agent-independent perspective, the game would look near-random due to its zero-sum nature, and the entropy would be 1.0 for all effective purposes.

%----- Polycraft --------
\subsection{Polycraft World (Action Domain)}
Minecraft is the best-selling video game of all time \cite{Minecraft}. It is a sandbox, open world video game that simulates earth-like 3D environments where players collect resources and blocks that can be used in a great variety of challenges. Polycraft World is a mod for Minecraft, originally developed to teach chemistry and materials science \cite{smaldone-natchem-2017} , which was extended to create the Polycraft World AI Lab (PAL) that generates complex open world scenarios for AI agents and human-AI teams \cite{goss2023polycraft}. The Polycraft World simulation environment adds thousands of additional items and inventories to typical Minecraft environments and provides an API that enables AI agents with diverse architectures to engage with the Minecraft world and undergo training and evaluation in various tasks. Polycraft World scenarios represent an interesting challenge for calculations of domain complexity as the environments can be arbitrarily large with thousands of interacting objects, providing a complex backdrop for AI tasks.

Environment space: The environment space in Polycraft World can vary based on task from small patches (e.g., 5 x 5 x 1 blocks) to arbitrarily large worlds that are effectively infinite (e.g., 1,000,000 x 1,000,000 x 256 blocks). The Polycraft World environment has numerous objects including thousands of resources (e.g., gold, wood, oil), thousands of tools (e.g., stone shovel, diamond axe, jet pack), and dozens of inventories for crafting and material processing (e.g., crafting table, chemical processor, distillation column). Objects can exist as blocks that can be mined, as floating entities that can be picked up, or as items stored in the agent’s inventory or another inventory such as a chest or a safe. The environment can also include goal-oriented agents (hereafter referred to as ‘actors’) other than the AI agent that perform actions that may be helpful, harmful, or neutral to the AI agent’s goals. Polycraft World tasks are typically turn-based, where the AI agent can plan in between turns without actors continuously modifying the environment. AI agents can obtain visual and symbolic information about the environment including the current game screen, their inventory, available recipes, all of the object, block, and actor locations in their current room, and the list of all past actions of all actors in their current room (Table \ref{tab:polycraftcommands}). The presence of actors with hidden goals and inventories and the inability to sense the interior of rooms the agent is not currently in creates realistic properties such as partial observability and non-determinism. 

Task solution space: The task solution space similarly varies based on task. To compare relatively efficient versus inefficient solutions, the performance score in a Polycraft World task is the sum of a positive reward received for completing the task (e.g., +256,000) and the negative costs associated with each action that are proportional to how resource-intensive the action is in the real world or for a human playing Minecraft (e.g., -100 for moving 1 block, -5,000 for chopping down a tree). While the task solution space can be large, it is typically much smaller than the environment space, particularly for a satisficing agent. For example, let’s imagine an AI agent with the goal of chopping down 1 tree who wakes up in a large forest arena with 1,000,000 trees and 1,000 objects that can be used through complex recipe paths to craft 100 different tools that all make chopping down a tree less expensive. If the tree and object types and locations are randomized, the number of potential environmental states is astronomical. Similarly, all trees, objects, and crafting paths are part of viable task solutions and create a large task solution space. Despite this complexity, the minimum viable task solution path is quite simple and can be accomplished by moving to the closest tree and using the BREAK\_BLOCK command.

\begin{figure}[h]
\centering
\includegraphics[width=15cm]{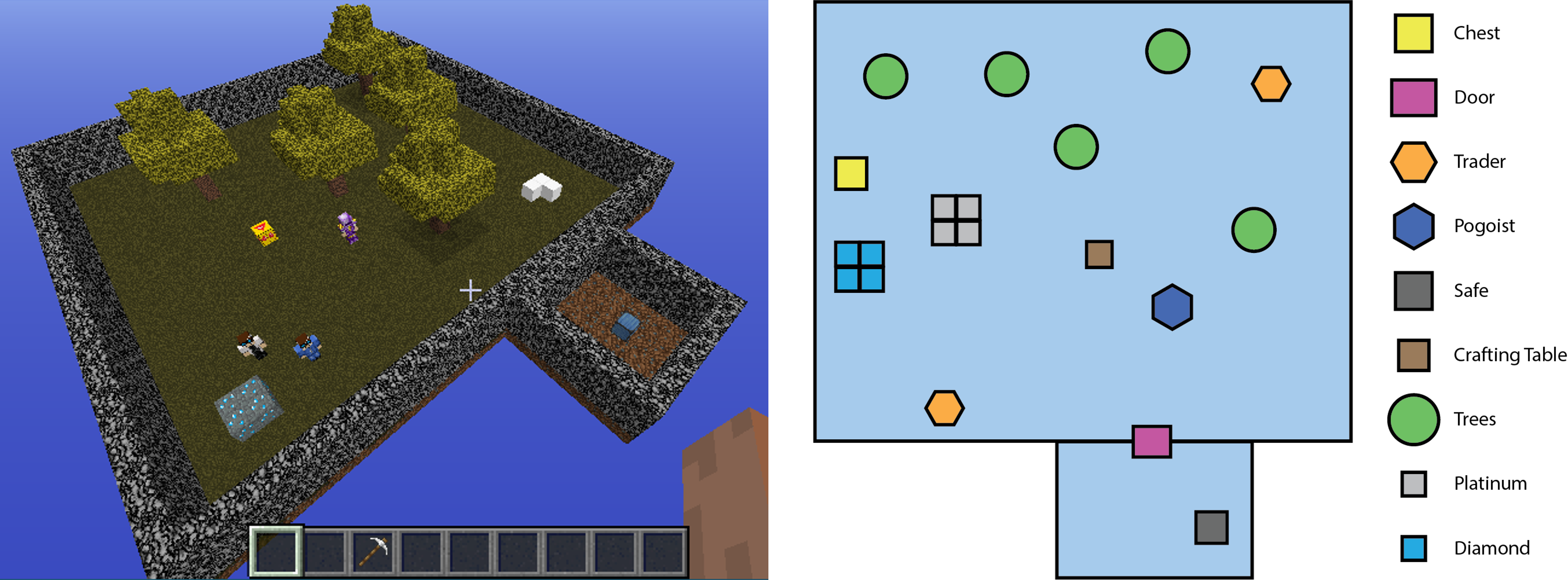}
\caption{POGO Task. Aerial in-game view (left) and schematic (right)}
\label{fig:POGO_task}
\end{figure}

Example task: Here we choose the example task POGO to illustrate complexity calculations for individual tasks in the Polycraft World domain (Figure \ref{fig:POGO_task}). The POGO task is a multi-step planning challenge in which the agent’s objective is to craft a pogo stick. The agent begins in a 32 x 32 x 3 grass field arena enclosed by unbreakable bedrock walls (the main room) with one or two side rooms measuring up to 16 x 16 blocks. At task start the arena includes 5 trees, 4 diamond blocks, 4 platinum blocks, 1 adversarial actor called the Pogoist, 2 allied actors called Traders, 1 crafting table, 1 chest with a key in it, up to 2 doors, and a safe in a side room that contains additional diamond. The agent’s inventory starts with 1 iron pickaxe that must be equipped to mine diamond and platinum ore. The Pogoist actor does not directly harm the AI agent but competes with the agent for resources. To accomplish the task, the agent must perform a mandatory series of steps including:

\begin{itemize}
    \item Mining resources (trees, diamond, platinum).
    \item Trading resources with Trader agents.
    \item Crafting intermediate objects, including a tree tap.
    \item Placing a tree tap on a tree to generate and collect rubber.
    \item Crafting a pogo stick.
\end{itemize} 

If the competing Pogoist actor depletes needed resources such as trees or diamond, additional steps may be needed including planting new trees from saplings that fall down from chopped trees, obtaining the key from the chest, unlocking doors, navigating to side rooms, and collecting resources from a safe. The task ends when the agent successfully crafts the pogo stick, gives up, or reaches the time limit.

\paragraph{\textbf{Dimensionality}}
As previously mentioned, the game complexity metrics have been designed for the analysis of static, discrete games, so their application to unmodified Minecraft or Polycraft World requires some adaptation. The POGO task was built in a discrete time context, where the game will only move one tick forward when an action is taken. We start with the intrinsic domain complexity (agent-independent) estimation where we want to answer ``How big is the problem space?''. For the POGO task, the map contains a main room with 30*30 available positions. One or two additional rooms may be present with up to 14*14 available positions. So, if POGO is considered as a 2D task, the number of possible positions is up to 1292 available positions. In addition to these, there are 36 inventory slots that can be occupied with items. The action space in Minecraft has been well studied \cite{MineDojo} and \cite{MineRL}; the available commands in the POGO task are listed in Table \ref{tab:polycraftcommands}.

\begin{center}
\begin{table}
\scriptsize
\caption{POGO Task Dimensionality}
\label{tab:pogo_dimensionality}
\begin{tabular}{|l|l|l|} \hline
{\bf Measure} & {\bf Description} & {\bf Value} \\ \hline
Arena size & Number of unique available positions in the arena & Up to $1292$ \\ \hline
Number of states & Number of world-state configurations reachable during the task & $1.2\times 10^{60}$ \\ \hline
State-space complexity & $Log_{10}(number of states)$ & $60.1$ \\ \hline
Maximum game length & Maximum allowed number of actions per task instance  & $2000$ \\ \hline
Average game length & Typical number of actions used by successful AI agents & $500$ \\ \hline
Branching factor & Number of possible actions & $43$ \\ \hline
Game-tree complexity & $Log_{10}(a^b)$; $a$: Branching factor; $b$: Average game length & $816.7$ \\ \hline
Game-space complexity & $Log_{10}(Number of states \times Maximum game length \times Branching factor)$ & $65.0$ \\ \hline
\end{tabular}
\end{table}
\end{center}

Counting the items in Table \ref{tab:polycraftcommands} is not sufficient to describe the number of distinct actions in the POGO task. To get a complete classification, we would have to consider the 36 possible slots in the inventory and the 5000 possible items available for the commands “PLACE”, “SELECT\_ITEM”, ”USE” and “DELETE”. The possibile combinations will change during the game as items availability changes. Further, if using the teleport command (“TP\_TO”), we would have to add to the previous number all available spaces in the maps (up to 1292). For simplicity, we don't consider the teleport (TP\_TO) command and, because the agents have their inventory empty most of the time, we will count the “PLACE”, “SELECT\_ITEM”, ”USE” and “DELETE” actions only once per potentially useful item in the task (Pickaxe, Tree Tap, Sapling, and Key). We don't count the sensing actions because they don't advance the game tick forward; however, we do account for the movement actions, “TURN” (right and left) and “TILT” because they do. We will consider that the “Move command” corresponds to 8 possible different actions (8 different directions). Taken together, these estimates give 43 possible actions that change the world state.

\setlength{\tabcolsep}{4pt} % Default value: 6pt
\begin{center}
\begin{table}
\scriptsize
\caption{Commands available in the POGO task}
\label{tab:polycraftcommands}

\begin{tabular}{lclclclc}
{\bf Interactions} & \# unique & {\bf Sensing} & \# unique & {\bf Movement} & \# unique & {\bf Game commands} & \# unique \\ \hline
BREAK\_BLOCK & 1 & SENSE\_SCREEN & 0 & MOVE & 8 & START & 0 \\
PLACE & 4 & SENSE\_ALL & 0 & TURN & 2 & RESET & 0 \\
COLLECT & 1 & SENSE\_ALL\_NONAV & 0 & TILT & 1 & GIVE\_UP & 1 \\
CRAFT & 6 & SENSE\_INVENTORY & 0 & TP\_TO (teleport to) & 0 & & \\
SELECT\_ITEM & 4 & SENSE\_LOCATIONS & 0 & & & &  \\
USE\_HAND & 1 & SENSE\_ACTOR\_ACTIONS & 0 & & & & \\
USE & 4 & SENSE\_RECIPES & 0 & & & &  \\
DELETE & 4 & SENSE\_ENTITIES & 0 & & & & \\
INTERACT & 1 & & & & & &\\
TRADE & 4 & & & & & &\\
NOP (no operation) & 1 & & & & & & \\
\end{tabular}
\end{table}
\end{center}

The total number of different states is determined by multiplying the number of different configurations and positions for each of the task elements. This gives a value of $1.2 \times 10^{60}$ and a state-space complexity of 60.1. Considering the number of possible actions gives a branching factor (43) and an average game length of 500 actions (average number of actions from evaluated AI agents), the value for Game-Tree Complexity is 816.7. This is probably a low estimate as the Branching factor will change every play. Multiplying the number of states, the maximum game length and the branching factor and taking the base 10 logaritm we obtain the game-space complexity of 65.0. All dimensionality values for POGO are shown in Table \ref{tab:pogo_dimensionality}. 

\begin{center}
\begin{table}
\scriptsize
\caption{Breakdown of the POGO task elements used in the calculation of diversity}
\label{tab:polycraft_task_elements}
\begin{tabular}{|l|c|c|c|} \hline
{\bf Task Element} & {\bf \# of entities} & {\bf Units of information} & {\bf $p_{i}$}\\ \hline
Diamond Ore & 4 & 12 & 0.017 \\ \hline
Platinum & 4 & 12 & 0.017 \\ \hline
Agent & 1 & 86 & 0.13 \\ \hline
Trees & 5 & 15 & 0.022 \\ \hline
Chest & 1 & 81 & 0.119 \\ \hline
Traders & 2 & 172 & 0.253 \\ \hline
Pogoist & 1 & 86 & 0.126 \\ \hline
Crafting Table & 1 & 81 & 0.119 \\ \hline
Safe & 1 & 81 & 0.119 \\ \hline
Doors & 2 & 6 & 0.009  \\ \hline
Branching factor & 43 & 43 & 0.070 \\ \hline

\end{tabular}
\end{table}
\end{center}

\paragraph{\textbf{Sparsity}}
The sparsity of the POGO task in Polycraft World is very close to zero. Calculating even an estimate of sparsity is not a trivial task due to the large number of elements, available actions and temporal relationships between steps. Let's define sparsity as the number of potentially successful sets of actions divided by the total sets of actions or game paths. We estimate the total number of game paths with the same assumptions as in the calculation of game-tree complexity: we consider an average game length of 500 plays and 43 possible actions per play. The approximate number of possible game paths is 43\textsuperscript{500} (on the order of  10\textsuperscript{816}). This is fairly straightforward. It is much more difficult to calculate the number of potentially successful sets of actions especially with agent movement. Even so, here, we perform a rough calculation of the sparsity of a drastically simplified POGO task to give a clear picture of the difficulties one may encounter.

Imagine a task scenario where the agent's goal is to collect wood from a tree. The agent starts 12 movements away from a tree (up - 7, left/up diagonal - 5). For simplicity, the agent will be facing the tree when next to it. To gather wood, the agent must perform 12 MOVE commands to get into position with the tree in front of it and then use the ‘BREAK\_BLOCK’ command.
A successful run for the agent is a permutation of the 12 MOVE commands: $(choose(12,7))=792$. Any additional MOVE commands may be played, however the opposite MOVE command must also be executed. These movements can be countered by one or more movements. Now, for additional simplicity, let’s set the maximum number of actions to 13. This way, no extra movements would allow for successful completion of the task. With these simplifications we will have a number of successful game paths of 792 and the possible game paths of 43\textsuperscript{13}. This would give a sparsity of $4.6 \times 10^{-19}$.

This value will only decrease for any extra movement allowed because the possibilities will always increase by 43 while the successful runs will always increase by an equal or smaller value. This means that 10\textsuperscript{-19} is an upper limit to the POGO task sparsity and the POGO task will only get more sparse as it grows in complexity. 
  
%\begin{center}
%\begin{table}
%\scriptsize
%\caption{Abundance of blocks and useful commands associated with each}
%\label{tab:polycraftblockabundance}
%\begin{tabular}{|l|l|l|} \hline
%{\bf Block Type} & {\bf Abundance} & {\bf Useful Commands} \\ \hline
%Air & 98.22\%, (1271/1294) & MOVE, TURN \\ \hline
%Trees & 0.39\%, (5/1294) & BREAK\_BLOCK, PLACE, TURN \\ \hline
%Diamond Ore & 0.31\% (4/1294) & BREAK\_BLOCK, TURN \\ \hline
%Platinum & 0.31\% (4/1294) & BREAK\_BLOCK, TURN \\ \hline
%Door & 0.15\% (2/1294) & USE, USE\_HAND, TURN \\ \hline
%Traders & 0.15\% (2/1294) & TRADE, INTERACT, TURN \\ \hline
%Pogoist & 0.08\% (1/1294) & TURN \\ \hline
%Crafting Table & 0.08\% (1/1294) & CRAFT, TURN \\ \hline
%Chest & 0.08\% (1/1294) & COLLECT, TURN \\ \hline
%Safe & 0.08\% (1/1294) & USE, COLLECT, TURN \\ \hline
%Tree Tap & 0.08\% (1/1294) & COLLECT, TURN \\ \hline
%\end{tabular}
%\end{table}
%\end{center}

\paragraph{\textbf{Diversity}}
For characterization of domain complexity, it is natural that a detailed description of the domain should be the main focus and the primary material for any calculations. In this spirit, we decided to calculate the Shannon entropy as a diversity index using a description of the POGO task based not on potential world state permutations, but on the minimal units of information needed to define a typical world state at a given time.

The steps to implement this method are the following. We start with a description of the task, one that is as thorough as possible. We identify the different elements that contribute to the description / complexity of the domain using the components described in section \ref{sec:Components}. We next calculate how many units of information are required to define each element present in the domain as follows. For objects without inventories (e.g., Diamond ore, Platinum, Trees, Crafting Table, and Doors) we assume that 3 units of information are needed to specify each instance of the entity, which are the name of the entity and the x and z coordinates (with the y coordinate being uniform in the 2D simplification of POGO). For objects with inventories (e.g., Chest and Safe) we assume that 81 units of information are needed to specify each instance of the entity, which are the name of the entity, x and z coordinates, and the object and number of objects in each of 39 inventory slots. For any agent with an inventory (e.g., the AI agent, Pogoist, and Traders) we assume that 86 units of information are needed to specify each instance of the entity, which are the 81 units defined above for inventories plus the 4 pieces of equipment each agent is wearing and the 1 item currently being held by the agent. The values in Table~\ref{tab:polycraft_task_elements} are calculated as the above units of information needed per entity multiplied by the number of each entity in a typical task environment.

We transform these values to create a percentage of the POGO domain’s complexity defining components that will be used as the $p_{i}$ in the normalized Shannon entropy, equation \ref{eq:EntropyNormalized} as showed in section \ref{sec:diversity}. We obtained a normalized entropy value of 0.870 for the POGO task.

The advantage of this method for calculating entropy is that it naturally follows from the definition of entropy: which can be defined as the amount of information needed to characterize an entity \cite{Shannon_paper}. We are using the description of the task to calculate the entropy. The main disadvantage is that there is no strict instruction set to create a standardized characterization of a domain. The main guideline is that the characterization should be sufficient to describe the domain, which we attempted to describe in section \ref{sec:Components}. This still leaves a large margin for interpretation, so this characterization is subjective. Despite this, there are some procedures that can mitigate the subjectivity, namely: check if the elements with the highest percentages can be subdivided and if that will improved description of the domain and check if each element's percentage aligns with the element's contribution to the domain's complexity.    

%----- MNIST --------
\subsection{MNIST (Perception Domain)}
MNIST is a dataset of handwritten arabic numbers from 0 to 9. The dataset contains cropped grayscale images of handwritten numbers. There are 60,000 training images and 10,000 testing images. Each image is 28x28 pixels. (\cite{colah}) This dataset is often used to train a classification algorithm, with the goal being to accurately recognize the digits regardless of handwriting style.
\paragraph{\textbf{Dimensionality}}
Each image is 28x28 pixels, and is evaluated as an array of the same dimensions. Each value in the array, for the purposes of this study, is a binary value of either 1 or 0. A value of 1 indicates that the pixel is part of the handwritten digit, while a value of 0 indicates background space. Each array is flattened into a $28*28 = 784$ vector. With $70,000$ total images, the dimensionality of the entire MNIST dataset can be considered a $70,000 * 784$ pixel array  times the 10 classes and times 2 as each image has binary values (\cite{colah}). When we take the ten based logarithm of this number (Equation \ref{eq:imageDim}) we get 9 for the dimensionality of the MNIST domain. 

\begin{figure}[htb]
  \centering
  \includegraphics[width=14cm]{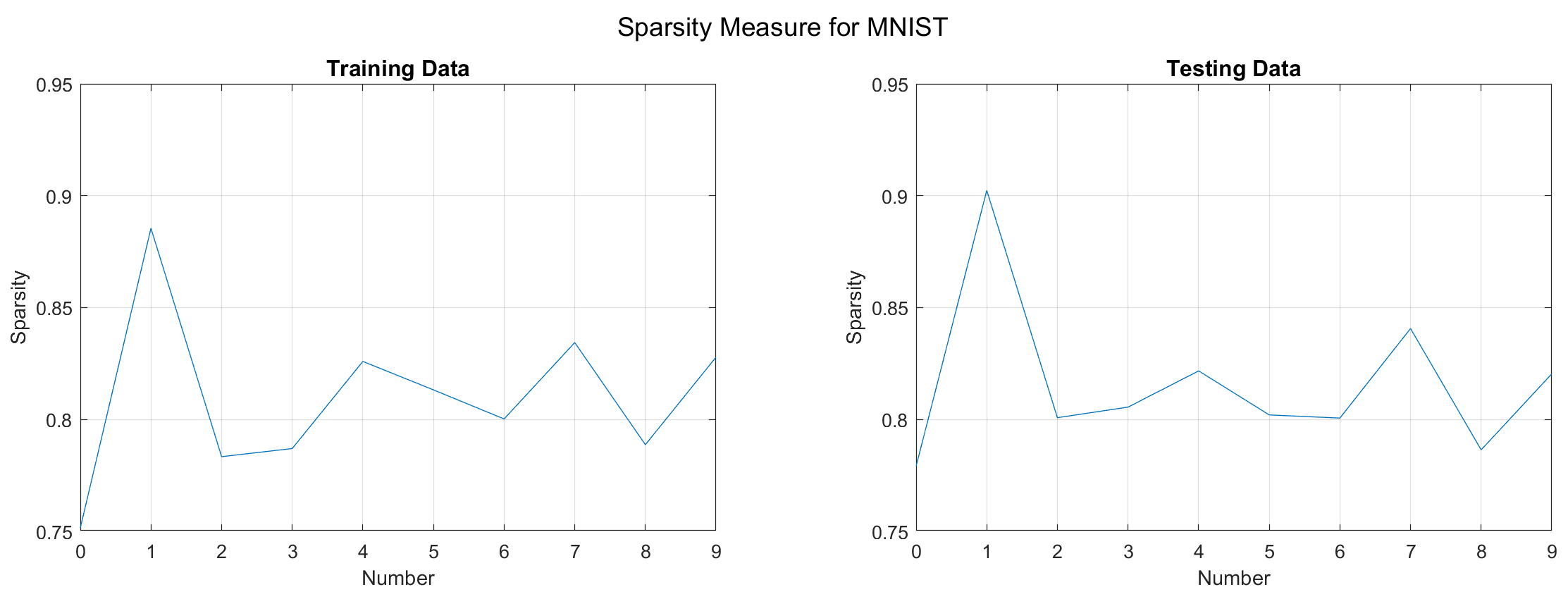}
  \caption{The average sparsity value for each digit in MNIST dataset. The left graph shows the average sparsity values of the training data and the graph on the right side of the testing data of the MNIST dataset. The sparsity values range from the lowest of 0.751, to the highest of 0.9, with average for the whole MNIST dataset being 0.813.} \label{fig:MNIST_sparsity}
\end{figure}

\paragraph{\textbf{Sparsity}}
Since this study uses binary value arrays, it is possible to calculate the sparsity of an image simply by taking the proportion of 1s within a pixel matrix. The formula below was used:
% formula for sparsity
\begin{align}
    Sparsity = \frac{nonzero\; elements\; of\; X}{all\; elements\; of\; X}
    \label{eq:MNIST_sparsity}
\end{align}
where X is the matrix of an image. 
%The majority of the image has little to no useful information. In other words, the majority of each image is ``background'' instead of the handwritten digit. Some images from the MNIST dataset have a lower or higher sparsity than the 80\%. 
Figure \ref{fig:MNIST_sparsity} shows the sparsity values for each digit. The sparsity values of the training and the testing data of the MNIST dataset are similar. The sparsity values range from the lowest of 0.751, to the highest of 0.9. The average for the whole MNIST dataset is 0.813, which reflects the high sparsity level of the dataset, meaning that there is lots of background, useless pixels in each image.  This is true especially for the digit ``1'' as shown in equation \ref{fig:MNIST_sparsity}.

\begin{figure}[htb]
  \centering
  \includegraphics[width=14cm]{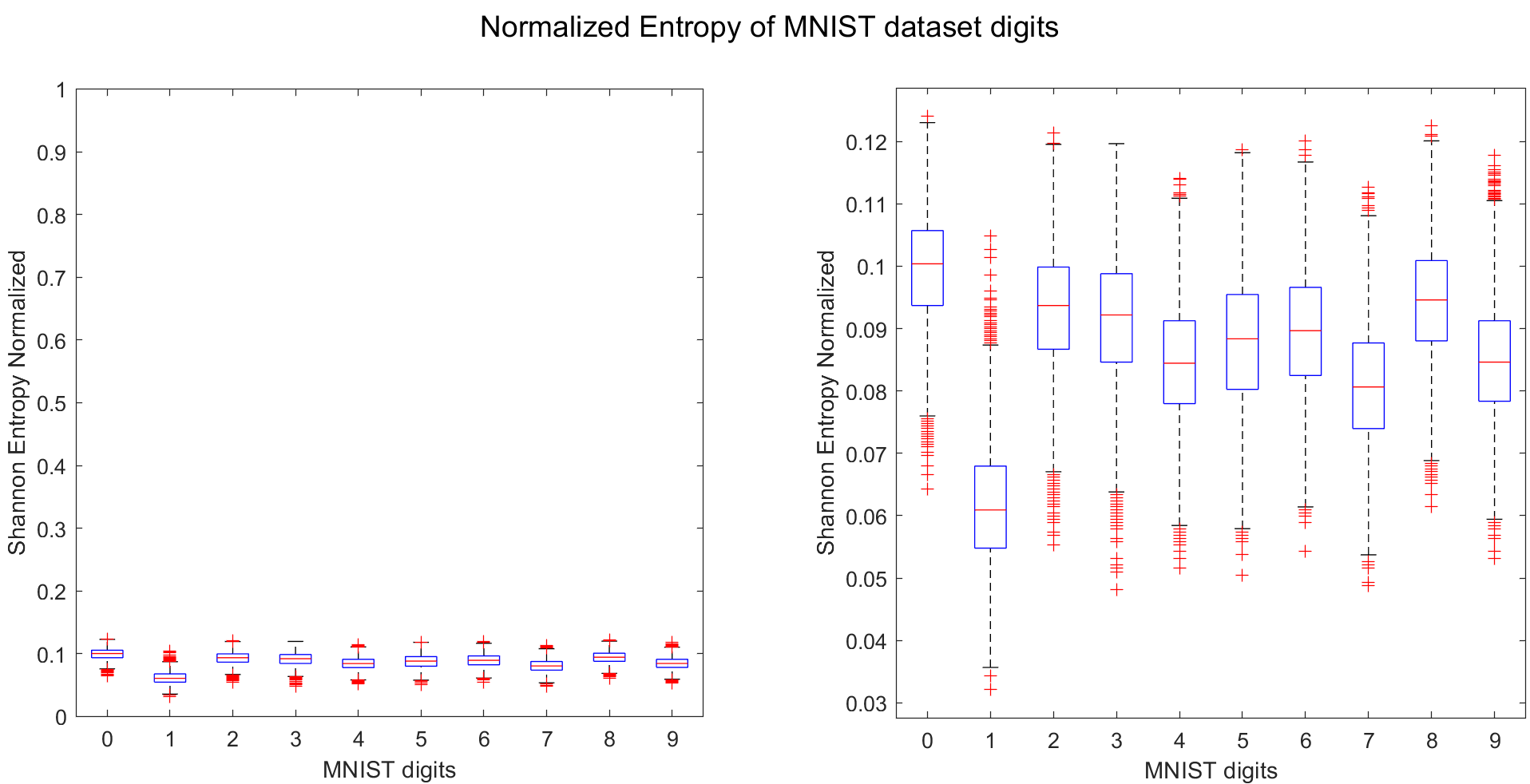}
  \caption{Normalized entropy of each digit in the MMNIST dataset. The boxplot on the left is the full range [0 1] of the normalized entropy values, whereas the boxplot on the right is zoomed to the results for better visibility. 
  The red line in the blue boxes represents the median of the entropy distribution for each digit, with the lowest value of 0.06 for digit ``1'' and the highest value of 0.1 for digit ``0''. All ten digits' entropy has normal distribution. The red whiskers show the outliers. Overall the entropy values are low for the training part of the MNIST dataset, which is 60000 images.}  
  \label{fig:entropy_MNIST}
\end{figure}

\paragraph{\textbf{Diversity}}
We calculated the Normalized Shannon Entropy for each image of each digit separately in the MNIST train dataset, to get possible range [0 1] for the entropy. We used equation \ref{eq:EntropyNormalized} for the calculation with probability being of the intensity  value of each grayscale image, with 256 bins. Figure \ref{fig:entropy_MNIST} boxplots shows the range of entropy for each digit. All digits have low entropy values, with highest median entropy value  0.1 of the digit ``0'' and the lowest of 0.06 for the digit of ``1''. The median of median entropy is 0.090. The very low entropy shows the high skewness of the probability distribution used for its calculation, since most of the bins are empty and all the pixels intensity are or in the first (black) or the last bin (white). Therefore, the very low entropy of MNIST dataset shows low diversity, and low uncertainty. 

%The MNIST dataset was divided into smaller datasets by class label, and within each dataset, every row represented an image. Each row's entropy value was calculated using the Shannon entropy function from Python's SciPy library. The log base 10 was taken for each value calculated. %(\cite{scipy_entropy})

%Most of the digits have an average entropy value ranging from 0.5-0.6, except for the 1 digit, which has a lower average of approximately 0.3. Across the distributions, there are extreme outliers, but each digit has a sample of 5000-6000 values, so the small number of outliers are likely numbers that were written in a smaller size, or a drastically different style from what is considered the norm.

%----- Cifar-10 --------
\subsection{CIFAR-10 (Perception Domain)}
The CIFAR-10 dataset contains 60,000 colour images. The images have ten classes. There are 50,000 training and 10,000 test images in the dataset. Each image is 32 x 32 pixel size and each pixel has a corresponding value for red, green and blue channels (or bands). There are ten classes: airplane, automobile, bird, cat, deer, dog, frog, horse, ship, and truck. Each class has 6,000 images (\cite{cifar10_info}).

\paragraph{\textbf{Dimensionality}}

Each image in the CIFAR-10 dataset is 32x32 pixels, with three color channels (or bands): red, green, and blue (RGB). Each pixel values are from 0 to 256, that is the intensity of the pixel. There are 60,000 images total, and ten class labels. Using Equation \ref{eq:imageDim}:
\begin{equation}
    D_{upperbound} = log_{10}(32 * 32 * 3 * 10 * 256 * 60000) = 11.67
\end{equation}
Thus, the CIFAR-10 dataset domain dimensionality is  11.67.

\begin{figure}[htb]
  \centering
  \includegraphics[width=15cm]{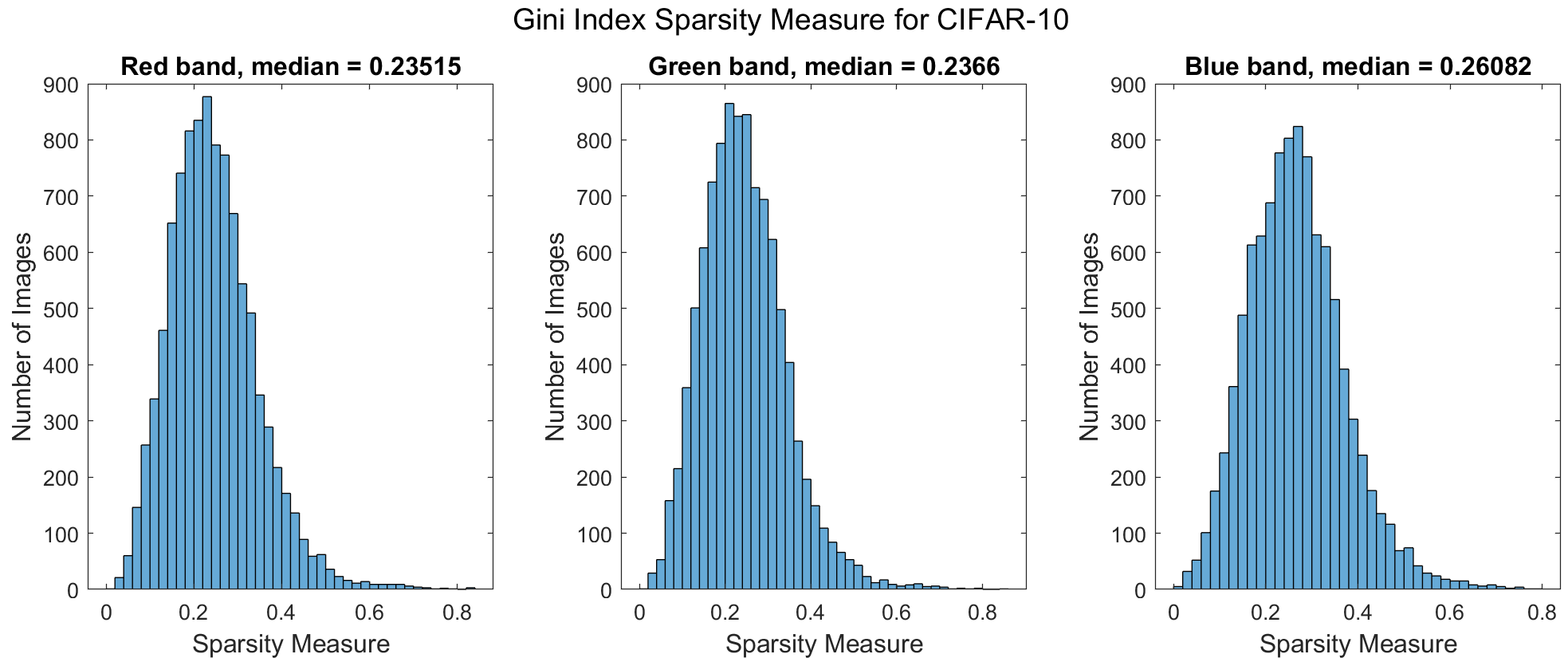}
  \caption{ The sparsity histogram of the CIFAR-10 data set using the \emph{Gini Index} from equation \ref{eq:Gini}. From left to right the graphs are for the red, green, and blue channels. The x-axis shows the sparsity measure from the \emph{Gini Index}, which spans from 0 to 1. The y-axis shows the number of images being used. } \label{fig:CIFARsparsity_measures}
\end{figure}

\paragraph{\textbf{Sparsity}}
We used the \emph{Gini Index}, from equation \ref{eq:Gini} to obtain the sparsity level of the CIFAR-10 dataset. The \emph{Gini Index} is capable of handling null values, which is required for this case. Figure \ref{fig:CIFARsparsity_measures} 
%and \ref{fig:CIFARsparsity_variations} 
shows the results of the \emph{Gini Index} calculation for each RGB color band. The distribution of sparsity values are Gaussian and similar with median values for sparsity for red: 0.235, green: 0.237 and blue: 0.26. The median of the median sparsity of the CIFAR-10 dataset is 0.237.  

%\begin{figure}[htb]
%  \centering
%  \includegraphics[width=15cm]{figures/sparsity_variations_cifar-10.png}
%  \caption{ These three graphs show the sparsity measure for the CIFAR-10 data set. From left to right the graphs are for the red, green, and blue channels of the images in the data set. On x-axis we can see the image number and the y-axis shows the sparsity level in the range of 0 to 1. } \label{fig:CIFARsparsity_variations}
%\end{figure}

\begin{figure}[htb]
  \centering
  \includegraphics[width=14cm]{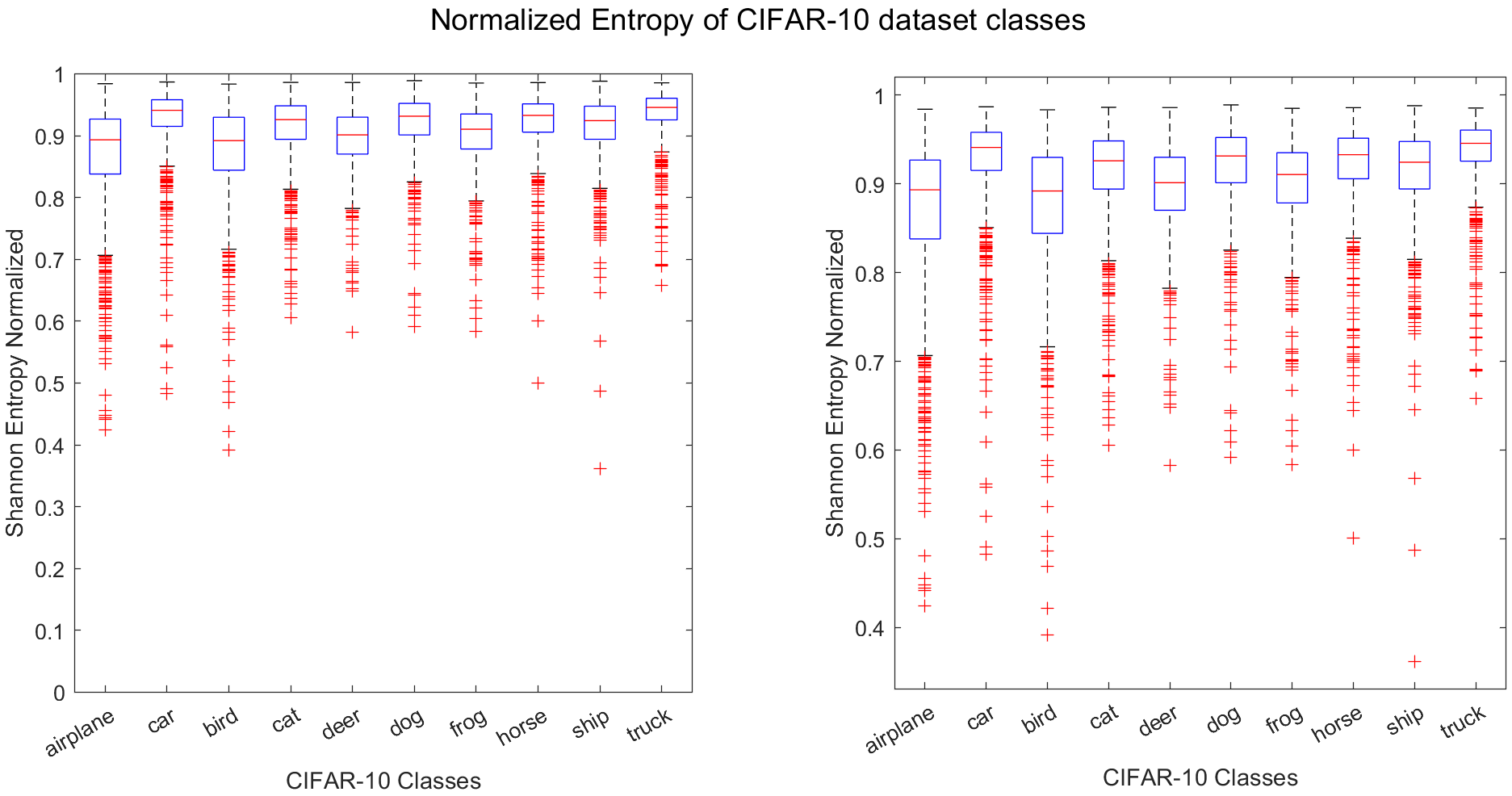}
  \caption{Normalized entropy of each class in the CIFAR-10 dataset. 
  The boxplot on the left is the full range [0 1] of the normalized entropy values, whereas the boxplot on the right is zoomed to the results for better visibility. 
  The red line in the blue boxes represents the median of the entropy distribution for each digit, with the lowest value of 0.892 for ``bird'' class and the highest value of 0.946 for for the ``truck'' class. The entropy distribution of each class varies from normal to skewed. There are many outliers toward the lower entropy values (red whiskers). Overall the entropy values are high for the CIFAR-10 dataset.}  
  \label{fig:entropy_CIFAR10}
\end{figure}

\paragraph{\textbf{Diversity}}
We calculated the Normalized Shannon Entropy for each image separately for each of the ten classes of the CIFAR-10 dataset, to get the possible range [0 1] for the entropy. We used equation \ref{eq:EntropyNormalized} for the calculation with probability being of the intensity value of each RGB band as grayscale image. That is 256 bins. Figure \ref{fig:entropy_CIFAR10} boxplots shows the range of entropy for each of ten classes. All classes have high entropy values, with the lowest value of 0.892 for ``bird'' class and the highest value of 0.946 for for the ``truck'' class. The entropy distribution of each class varies from normal to skewed. There are many outliers toward the lower entropy values (red whiskers). Overall the entropy values are high for the CIFAR-10 dataset.
The median of median entropy is 0.925. This high entropy value that is close to the highest eveness of the domain, reflects low uncertainty and shows low diversity of the domain.

%----- Iris --------
\subsection{Iris (Data Science Domain)}
The Iris dataset is a multivariate data set created by Ronald Fisher in his 1936 paper \emph{The Use of Multiple Measurements in Taxonomic Problems}. The dataset contains 3 classes of iris flower - Iris Setosa, Iris Versicolour, and Iris Virginica - with 50 instances of each. Each instance consists of four different measurements of an Iris flower: petal length, petal width, sepal length, and sepal width. (\cite{https://doi.org/10.1111/j.1469-1809.1936.tb02137.x})

\paragraph{\textbf{Dimensionality}}
The iris dataset is a 150x5 dimensional dataset. The attributes are as follows: petal length, petal width, sepal length, and sepal width. The four measurements are continuous (numerical) data. Each row of data has a corresponding class label in form of categorical data.

\paragraph{\textbf{Sparsity}}
The Gini index (see Equation \ref{eq:Gini}) was used to calculate the sparsity of each class and each attribute in the Iris data set. Table \ref{fig:Iris Sparsity} shows the results of the sparsity calculation. Overall we see lower sparsity levels. 

\begin{table}[h]
\begin{tabular}{|c|c|c|c|}
\hline
      \textbf{Attributes} & \textbf{Setosa} & \textbf{Versicolor} & \textbf{Virginica} \\
     \hline
     \textbf{Sepal Length} & 0.0392 & 0.0489 & 0.0533 \\
     \hline
     \textbf{Sepal Width} & 0.0602 & 0.0632 & 0.0589 \\
     \hline
     \textbf{Petal Length} & 0.0634 & 0.0610 & 0.0551 \\
     \hline
     \textbf{Petal Width} & 0.2086 & 0.0826 & 0.0759 \\
     \hline
\end{tabular}
\caption{This table shows the sparsity levels of each class and each attribute within the Iris data set. The classes are different types of irises that are represented. 
The sparsity value ranges from 0 to 1 where zero is least sparse and one is the most sparse.}  \label{fig:Iris Sparsity}
\end{table}

\subsection{Discussion} 
The six case studies presented in this section suggest that the three measures of dimensionality, sparsity and diversity can be applied broadly. They also allow for the incorporation of domain expertise in their computation, as they are systematic guidelines (and in many cases, guided by prior practice in that community) rather than hardened rules that do not allow for diversity of interpretation. For example, in the Monopoly domain, as in other multi-player games, computing dimensionality tends to be relatively straightforward and well-defined. It is also evident that the entropy of Monopoly is very high (close to 1.0), which suffices for practical applications. We cite this example to show that, even when exact values are computationally intractable to derive, certain assumptions can be made, which allow us to derive estimates of the true underlying complexity with high confidence. In another example, the Polycraft World domain illustrates interesting considerations for calculating complexity in large-scale 3D environments with many object types. Specifically, even in worlds where the number of potential environment states is astronomical due to scale, the amount of information needed to define a specific instance of the world state can still be quite small. 

Similar to other domains, the CartPole domains measure their dimensionality based on an enumeration of the dimensions of the state space and their scales. In terms of sparsity, the CartPole domains allow for a more analytical approach based on the ratio of successful action sequences to the total possible action sequences, which are both computable in CartPole, but not in more complicated domains like Monopoly or Polycraft World. The diversity measure is where CartPole deviates most from the other domains. CartPole uses an entropy-based diversity measure like other domains, but the entropy is calculated based on observable sequences of features and actions, not the possible sequences. This calculation does require interaction with the game, but the interaction can be random and not involve a particular agent, thus preserving the intrinsic property of the diversity measure. Combined with the approaches taken in other domains, the overall framework for computing dimensionality, sparsity, and diversity applies to action-based and classification-based domains, and in cases where either an analytical or empirical approach is warranted.

Ultimately, deriving values for such measures in a systematic way enables a robust and rich set of comparisons. To take just one example, if we compare MNIST and CIFAR-10, the MNIST dataset has very low normalized entropy, whereas CIFAR-10 has very high normalized entropy (the former is skewed, whereas the latter is more similar to the uniform distribution). For both cases, however, the entropy reflects both low diversity and low uncertainty, because the highest diversity is achieved when the normalized entropy is close to 0.5. 

\section{Conclusion}
\label{Conclusions}
Understanding the complexity of domains in a systematic and domain-independent manner is an important problem in the Artificial Intelligence community. Understanding domain complexity is especially relevant today, when the focus has shifted from building task-specific AI designed for rigidly defined problem, to the development of OWL agents and algorithms that can work across domains, or that are able to handle `novelties' and other unexpected events in a single open-world domain. Returning to our example of the AUV from \citeauthor{Wilson2014} (2014), we can express the difficulty that was encountered when transitioning from the simulator domain to the real-world domain in terms of the components of our complexity levels. The real world had higher dimensionality with regard to the environment space complexity and greater diversity with respect to the perception-task solution space (where noise and environmental factors meant a diverse collection of different sensor readings might represent a single system state). These unexpected increases in domain complexity, as compared to the simulator domain, caused the AI to behave pathologically, constantly replanning in response to the discrepancies between its perception of the real world and its less complex internal state space. Once the complexity of the deployment domain was understood, a bounding-box technique was introduced, enabling the robot to operate successfully in the open ocean. 

The goal of this article was to present a blueprint, and set of implementable measures, for systematically understanding domain complexity. We argued that domain complexity, especially in the context of AI-relevant domains, can be viewed from two aspects: intrinsic and extrinsic. We also proposed three categories of measures (dimensionality, sparsity and diversity) and applied them to a broad range of case studies across action, perception and data science domains. 

As the systematic understanding of domain complexity remains a relatively new problem in AI, many important questions (both empirical and theoretical) remain to be answered by future work. In the next section, we close the article by enumerating several promising questions.

%Wrapping up Mark's parable.  When the robot went from sim
% How would Mark benefited if had the knowledge of the complexity level of the domain of the simulation and the domain of the open-world  using the components and estimation we described in section 5 and 6? 

%=====================================================
\section{Next Steps}
This paper is an early attempt at synthesizing several measures and aspects of domain complexity that are hypothesized to apply generally across domains and tasks. Many questions remain. We believe that these questions need to be formulated and addressed within a research agenda that treats domain-\emph{general} complexity (as opposed to domain-\emph{specific} complexity) as a first-class citizen. We are not claiming that each such research question should apply empirically to every domain, but ideally, it should be applicable to a diverse set of domains, allowing general claims to be made. 
As next steps, we state three questions below that (potentially) could be investigated in an experimental setting with the goal of yielding general insights about domain complexity:

\begin{enumerate}
    \item Can we derive strong theoretical connections between domain complexity and \emph{difficulty}? For instance, will it always be the case that tasks in a more complex domain \emph{necessarily} will be more difficult, or just more difficult \emph{in expectation}? And can difficulty be studied independently of complexity? These are questions with which multiple communities, especially within AI and open-world learning, are only just beginning to grapple. In addition to stating theoretical claims, it is also important to test these claims empirically. We believe that designing appropriate experimental methodologies that allow us to validate general claims is, in itself, a promising area for future research.
    \item Can our measures of domain complexity be used to define and \emph{quantify} complexity in so-called complex systems, such as networks and dynamical systems, as well as other nonlinear systems (e.g., differential equations, computational fluid dynamics simulations, weather forecasting)? Is one system more complex than another, and if so, then along what perspectives (components)? Moreover, does this have theoretical ramifications, validated by appropriate experiments, for analyzing such systems? 
    \item In the cases of infinite action and state spaces, what are the appropriate mathematical frameworks for distinguishing between domains of (arguably) differing complexity? In real analysis, for instance, there are different hierarchies of infinity that are well understood. For instance, integers and real numbers both form infinite sets, but the latter has provably greater cardinality than the former. Could similar claims be made for complexity?
\end{enumerate}

These questions are not exhaustive, and some have several other questions associated with them that may require several parallel lines of theoretical and experimental research to investigate fully. We emphasize that the one commonality between all these questions is their lack of dependence on a single domain or model. Rather, all of them aim toward an agenda that is as domain-independent as possible. As noted earlier, we believe that this is the central element that distinguishes other field-specific studies of complexity from our proposal. 

%[making experiment: applying these measures to several domains and compare them]
%\subsection{Setup for experiments}

%\subsection{Future work}
%Expanding this framework to complex systems. 

\section{Acknowledgments}
This research was  sponsored  by DARPA and the Army Research Office (ARO) under multiple contracts / agreements, including  
% Eric, UTD contract#, 
W911NF2020010, 
% Mayank, USC-ISI contract#, 
W911NF2020003, and  
% Larry, WSU contract#.
W911NF-20-2-0004.
The views contained in this document are those of the authors and should not be interpreted as representing the official policies, either expressed or implied, of DARPA, NRL, ARO, or the U.S. government.

The following teams and individuals contributed to the open-world novelty hierarchy presented in Table 1: Washington State University led by Lawrence Holder, Australian National University led by Jochen Renz, University of Southern California led by Mayank Kejriwal, University of Maryland led by Abhniav Shrivastava, University of Texas at Dallas led by Eric Kildebeck, University of Massachusetts at Amherst led by David Jensen, Tufts University led by Matthias Scheutz, Rutgers led by Patrick Shafto, Georgia Tech led by Mark Riedl, PAR Government led by Eric Robertson, SRI International led by Giedrius Burachas, Charles River Analytics led by Bryan Loyall, Xerox PARC led by Shiwali Mohan, Smart Information Flow Technologies led by David Musliner, Raytheon BBN Technologies led by Bill Ferguson, Kitware led by Anthony Hoogs, Tom Dietterich, Marshall Brinn, and Jivko Sinapov.

We are especially grateful to Christine Task for collaborating, providing interesting ideas and extensive literature review for the initial version of this paper presented at the AAAI Spring 2022 Workshop on Designing Artificial Intelligence for Open Worlds \cite{Doctor2022}. We thank Russell Leong, Roman Sanelli, and Sarah DaSilva for conducting a literature review, proofreading, and providing edits of the paper. We especially thank  Daniel Smith for expanding the literature and sparsity calculations for MNIST and CIFAR-10 domains. The authors would also like to thank David Aha, Robert Lazar, and Ira Moskowitz for their feedback on this or previous version of this article.

%=====================================================
%=====================================================
%=====================================================
%=====================================================

\printcredits

%\nocite{*}
\bibliographystyle{cas-model2-names}

%Test \cite{Wilson2014}.

\bibliography{bibliography.bib}



\end{document}